%% file: nips2026_conference.tex
\documentclass{article}
\PassOptionsToPackage{numbers,compress}{natbib}
\usepackage[preprint]{neurips_2026}

\usepackage[utf8]{inputenc}
\usepackage[T1]{fontenc}
\usepackage{microtype}
\usepackage{hyperref}
\usepackage{url}
\usepackage{graphicx}
\usepackage{wrapfig}
\usepackage{pifont}
\usepackage{tikz}
\usepackage{pgf-pie}
\usepackage[table,dvipsnames]{xcolor}
\usepackage{booktabs}
\usepackage{longtable}
\usepackage{subcaption}
\usepackage{amsmath}
\usepackage{amsfonts}
\usepackage{nicefrac}
\usepackage{array}
\usepackage{tabularx}
\usepackage{enumitem}
\usepackage[most]{tcolorbox}

\definecolor{darkblue}{rgb}{0, 0, 0.5}
\definecolor{jbblue}{HTML}{355070}
\definecolor{jbgreen}{HTML}{6F9074}
\definecolor{jborange}{HTML}{BC6C25}
\definecolor{jbbrick}{HTML}{9C4330}
\definecolor{jbpeach}{HTML}{E8D5C4}
\definecolor{jbsand}{HTML}{F4EFE3}
\definecolor{jbmint}{HTML}{EEF4EA}
\definecolor{jbice}{HTML}{EDF3F6}
\definecolor{badgeanthropic}{HTML}{C96A28}
\definecolor{badgeopenai}{HTML}{0F766E}
\definecolor{badgegoogle}{HTML}{2563EB}
\definecolor{badgeqwen}{HTML}{1D4ED8}
\definecolor{badgedeepseek}{HTML}{0369A1}
\definecolor{badgemoonshot}{HTML}{BE123C}
\definecolor{badgeminimax}{HTML}{4338CA}
\definecolor{badgexai}{HTML}{374151}
\definecolor{badgeopenclaw}{HTML}{7C2D12}
\definecolor{badgeopencode}{HTML}{1D4ED8}
\definecolor{badgeother}{HTML}{6B7280}
\definecolor{domainops}{HTML}{31525B}
\definecolor{domainfinance}{HTML}{5C7457}
\definecolor{domainit}{HTML}{1F5673}
\definecolor{domaineng}{HTML}{7C5C3B}
\definecolor{domainsci}{HTML}{5B6C8D}
\definecolor{domainmedia}{HTML}{A05C4B}
\definecolor{domainlegal}{HTML}{7F4F24}
\definecolor{domainhealth}{HTML}{8C6A9A}
\definecolor{domainadmin}{HTML}{847F67}
\definecolor{pieBFO}{HTML}{2F4A62}
\definecolor{pieOffice}{HTML}{0D6B5F}
\definecolor{pieComputer}{HTML}{1B5A73}
\definecolor{pieEngineering}{HTML}{A46324}
\definecolor{pieManagement}{HTML}{8B4A21}
\definecolor{pieArts}{HTML}{934635}
\definecolor{pieSales}{HTML}{536F4A}
\definecolor{pieLegal}{HTML}{6E3B1D}
\definecolor{pieScience}{HTML}{4D5F82}
\definecolor{pieEducation}{HTML}{5F6873}
\newcolumntype{Y}{>{\raggedright\arraybackslash}X}
\newcommand{\cell}[2]{\shortstack[c]{\strut #1\\\strut #2}}
\newcommand{\leaderlogo}[1]{\raisebox{-0.22ex}{\includegraphics[height=1.75ex]{#1}}}

\tcbset{
  jbpromptstyle/.style={
    enhanced,
    breakable,
    colback=white,
    colframe=jbblue!80!black,
    colbacktitle=jbblue!10,
    coltitle=jbblue!85!black,
    fonttitle=\bfseries,
    boxrule=0.45pt,
    arc=1pt,
    left=5pt,
    right=5pt,
    top=5pt,
    bottom=5pt,
    before skip=0.75em,
    after skip=0.75em
  }
}
\newtcolorbox{promptbox}[1][]{
  jbpromptstyle,
  title=Prompt,
  #1
}
\newtcblisting{promptlisting}[1][]{
  jbpromptstyle,
  title=Prompt,
  listing only,
  listing options={
    basicstyle=\ttfamily\small,
    breaklines=true,
    columns=fullflexible,
    keepspaces=true,
    showstringspaces=false
  },
  #1
}
\newtcbinputlisting{\taskinstructionbox}[2][]{
  enhanced,
  breakable,
  colback=jbice!35,
  colframe=jbblue!85!black,
  colbacktitle=jbblue!12,
  coltitle=jbblue!85!black,
  fonttitle=\bfseries,
  boxrule=0.5pt,
  arc=1pt,
  listing only,
  listing file={#2},
  listing options={
    basicstyle=\ttfamily\scriptsize,
    breaklines=true,
    columns=fullflexible,
    keepspaces=true,
    showstringspaces=false
  },
  #1
}
\newtcbinputlisting{\taskrubricbox}[2][]{
  enhanced,
  breakable,
  colback=jbsand!45,
  colframe=jborange!90!black,
  colbacktitle=jborange!16,
  coltitle=jborange!80!black,
  fonttitle=\bfseries,
  boxrule=0.5pt,
  arc=1pt,
  listing only,
  listing file={#2},
  listing options={
    basicstyle=\ttfamily\scriptsize,
    breaklines=true,
    columns=fullflexible,
    keepspaces=true,
    showstringspaces=false
  },
  #1
}
\newtcbinputlisting{\taskcardbox}[2][]{
  enhanced,
  breakable,
  colback=jbmint!45,
  colframe=jbgreen!85!black,
  colbacktitle=jbgreen!16,
  coltitle=jbgreen!65!black,
  fonttitle=\bfseries,
  boxrule=0.5pt,
  arc=1pt,
  listing only,
  listing file={#2},
  listing options={
    basicstyle=\ttfamily\scriptsize,
    breaklines=true,
    columns=fullflexible,
    keepspaces=true,
    showstringspaces=false
  },
  #1
}
\newtcolorbox{taskrubricreadablebox}[1][]{
  enhanced,
  breakable,
  colback=jbsand!45,
  colframe=jborange!90!black,
  colbacktitle=jborange!16,
  coltitle=jborange!80!black,
  fonttitle=\bfseries,
  fontupper=\small,
  boxrule=0.5pt,
  arc=1pt,
  left=6pt,
  right=6pt,
  top=6pt,
  bottom=6pt,
  before skip=0.75em,
  after skip=0.75em,
  #1
}
\newtcolorbox{taskcardreadablebox}[1][]{
  enhanced,
  breakable,
  colback=jbmint!45,
  colframe=jbgreen!85!black,
  colbacktitle=jbgreen!16,
  coltitle=jbgreen!65!black,
  fonttitle=\bfseries,
  fontupper=\small,
  boxrule=0.5pt,
  arc=1pt,
  left=6pt,
  right=6pt,
  top=6pt,
  bottom=6pt,
  before skip=0.75em,
  after skip=0.75em,
  #1
}

\title{JobBench: Aligning Agent Work With Human Will}

\author{
\parbox{\textwidth}{\centering
\small
Yuetai Li$^{1}$\thanks{Equal contribution.} \quad Yichen Feng$^{1}$\footnotemark[1] \quad Zhangchen Xu$^{1,10}$ \quad Zixian Ma$^{1}$ \quad Kaiyuan Zheng$^{1}$ \\
Fengqing Jiang$^{1}$ \quad Xinghua Sun$^{1}$ \quad Rulin Shao$^{1}$ \quad Zichen Chen$^{2,3,10}$ \quad Yue Huang$^{6}$ \\
Xinyang Han$^{7}$ \quad Brian Lee$^{13}$ \quad Kayla Xu$^{5}$ \quad Shenglai Zeng$^{8}$ \quad Hang Hua$^{9}$ \\
Xiangliang Zhang$^{6}$ \hspace{0.6em} Basel Alomair$^{1,11}$ \hspace{0.6em} Ranjay Krishna$^{1}$ \hspace{0.6em} Luke Zettlemoyer$^{1}$ \hspace{0.6em} Pang Wei Koh$^{1}$ \\
Bhaskar Ramasubramanian$^{12}$ \hspace{0.6em} Luyao Niu$^{1}$ \hspace{0.6em} Xiang Yue$^{4}$ \hspace{0.6em} Radha Poovendran$^{1}$ \\[0.45em]
{\normalfont\small
$^{1}$University of Washington \quad
$^{2}$University of California, Santa Barbara \quad
$^{3}$Stanford University \\
$^{4}$Carnegie Mellon University \quad
$^{5}$Northwestern University \quad
$^{6}$University of Notre Dame \\
$^{7}$University of California, Berkeley \quad
$^{8}$Michigan State University \quad
$^{9}$MIT-IBM Watson AI Lab \\
$^{10}$Bake AI \quad
$^{11}$King Abdulaziz City for Science and Technology \quad
$^{12}$Western Washington University \quad
$^{13}$University of Chicago \\[0.35em]
{\scriptsize Homepage: \url{https://job-bench.github.io/}}\\[-0.1em]
{\scriptsize Hugging Face: \url{https://huggingface.co/datasets/JobBench/job-bench}}\\[-0.1em]
{\scriptsize GitHub: \url{https://github.com/Job-Bench/job-bench-eval}}}
}
}

\begin{document}

\maketitle

\begin{abstract}
Current benchmarks for occupational AI agents are scoped primarily by economic values, telling a replacement story. We introduce JobBench, which evaluates AI agents on the workflows that experts identify as high-priority for delegation, empowering humans based on their needs instead of replacing them with GDP value. JobBench covers 130 agentic tasks across 35 occupations. Each task is packaged as a workspace of heterogeneous reference files, requiring the agent to reason through the cluttered information streams of real professional work. Outputs are graded by a fact-anchored chain of rubrics, averaging 35.6 binary criteria per task. We evaluate 36 models; the strongest, Claude Opus~4.7 under Claude Code, reaches only 45.9 \%. We hope JobBench shifts the community's target labour-market effect from replacement to enhancement: building agents that do what humans actually want delegated, not only what is most economically valuable.
\end{abstract}

\input{sections/introduction}

\input{sections/jobbench_benchmark}
\input{sections/experiments}

\input{sections/conclusion}

\clearpage
\bibliographystyle{plainnat}
\bibliography{nips2026_conference}

\newpage

\appendix

\input{sections/appendix_toc}
\newpage

\input{sections/related_work}
\newpage

\input{sections/discussion}

\newpage

\input{sections/appendix_leaderboard}
\newpage

\input{analysis_outputs/jobbench_task_split_appendix}

\clearpage
\input{sections/appendix_task_examples}

\clearpage

\input{sections/appendix_prompts}

\newpage
\input{sections/appendix_supplementary_analysis}

\end{document}

%% file: sections/introduction.tex
\begin{figure*}[!ht]
\centering
\includegraphics[width=0.9\textwidth]{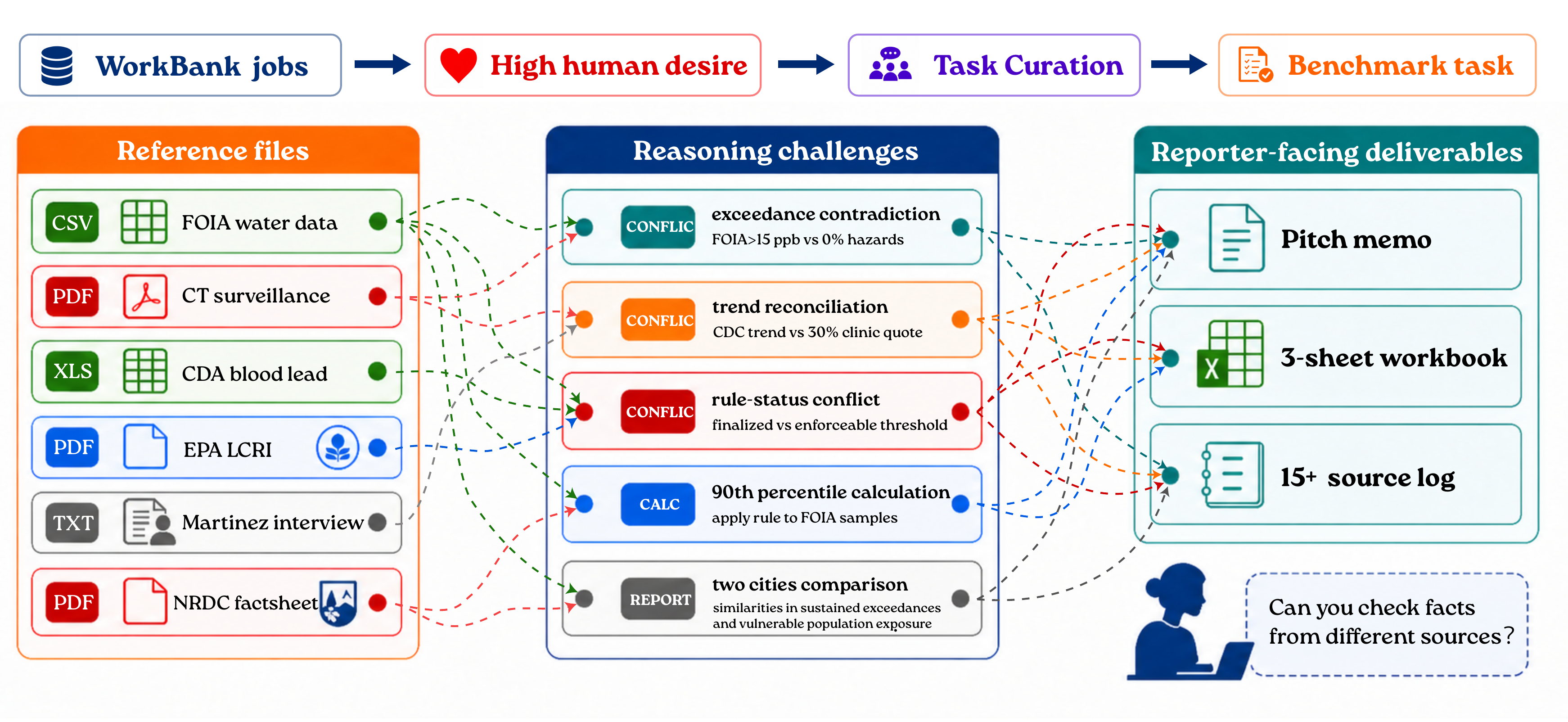}
\caption{JobBench overview. Tasks are constructed on top of Workbank \citep{shao2026futurework}, a survey in which over 1{,}500 workers report which of their work duties they would prefer AI to automate. We select 35 occupations at the intersection of high reported delegation preference and high economic exposure, then design 130 benchmark tasks aligned with expert will. The example illustrates a duty that reporters most want delegated to AI: ``checking different source reference materials to obtain relevant facts.'' Dashed lines trace each fact from its source file, through the reasoning challenge it raises, to the deliverable that fact ultimately supports.
}
\label{fig:jobbench:reporter_task_example}
\end{figure*}

\section{Introduction}
\label{sec:introduction}

The conversation about AI in the workplace has been framed almost entirely in economic terms: \textit{What fraction of working hours can agents absorb? How much of GDP is exposed to automation?} Workplace benchmarks inherit this framing by design. GDPVal \citep{openai2025gdpval} selects tasks that represent economic value and scores agents on whether they can deliver professional knowledge work judged against expert reference deliverables \citep{openai2025gdpval}. The Remote Labor Index measures end-to-end remote-work projects against contractor pay \citep{mazeika2025rli}.
\$OneMillion-Bench prices each of its 400 expert tasks by senior-expert hours times market wage and scores agents on the share of that expert-priced work they can reliably deliver \citep{yang2026onemillionbench}.
All these benchmarks ask the same question in different shapes: which economically valuable deliverables can an agent now produce on its own?

This framing, however, leaves a complementary criterion ignored. If agents are to share the professional workplace with the humans who hold those jobs, evaluations should weigh, which subset of its duties the workers themselves prefer to delegate. We treat this as a human-centered constraint on benchmark design: the professional is positioned not as labor to be displaced, but as a domain expert whose stated preferences over their own work inform which duties merit automation and enhance the productivity.

\textbf{JobBench} is a benchmark built on that principle. Every one of its 130 tasks across 35 professions is constructed from work that experts in that field claim they most want a capable agent to handle. We design tasks on top of Workbank \citep{shao2026futurework}, a worker-centered survey in which more than 1,500 workers rate, for every O*NET \footnote{The Occupational Information Network (O*NET), maintained by the U.S.~Department of Labor, provides a detailed list of work duties for each occupation.} work duty of their own occupation, whether they would prefer an AI agent to take that work over. We select the 35 occupations with high average willingness for automation, and develop the work duties into full benchmark evaluations. 
Figure~\ref{fig:jobbench:reporter_task_example} illustrates a reporter task example, grounded by a duty that reporters most want delegated to AI: ``checking different source reference materials to obtain relevant facts.'' Dashed lines trace each fact from its source file, through the reasoning challenge it raises, to the deliverable that fact ultimately supports.

JobBench grades each task with a chained rubric set that carries 35.6 binary criteria per task on average and 4,631 criteria in total. Every criterion is anchored to a deterministic number, a specific reasoning step, or a documented professional judgment, and a rubric receives its weight only when every criterion in the chain passes together. There is no partial credit for surfacing the right fact through a wrong inference. 

We summarize our contributions as follows:
\begin{itemize}
    \item \textbf{Align with human will.} We ground every JobBench task in domain expert delegation-desire, so that progress on the leaderboard maps onto how agents lift worker satisfaction and productivity together.
    \item \textbf{Professional reasoning.} Each task provides a workspace of heterogeneous reference files that may contain conflicts and hints for search. Credit goes to agents that retrieve and reconcile the right sources.
    \item \textbf{Fact-anchored chained rubrics.} The 4,631 binary criteria are organized into rubric chains that award credit only when every criterion in the chain passes. Pooled across sampled runs from different agents, 95.4\% of rubrics are passed at least once, evidence that each criterion is verifiable in practice.
    \item \textbf{Challenging for frontier models.} Across 36 agent configurations, the strongest setup, Claude Opus~4.7 under \texttt{Claude Code}, reaches 45.9\%; outside the Claude and GPT families no agent exceeds 19\%.
\end{itemize}

%% file: sections/jobbench_benchmark.tex
\section{The JobBench Benchmark}
\label{sec:benchmark}

\subsection{Design Principles of JobBench}
\label{subsec:Characteristics of JobBench}

\textbf{Align with human will.}
JobBench treats the expert's own willness and judgment as the selection signal, drawing on Workbank \citep{shao2026futurework}, in which more than 1,500 workers rate every work duty in their occupation for delegation desire. By targeting duties that experts want delegated and spend the most preparation time on, JobBench measures capability on work whose automation lifts worker satisfaction and productivity together. 

\textbf{From knowledge delivery to professional reasoning.} GDPVal \citep{openai2025gdpval} evaluates polished deliverables from relatively clean task packets. JobBench instead issues heterogeneous, sometimes conflicting workspaces where agents must locate, retrieve, and reconcile source evidence before producing the final artifact. This shifts the evaluation target from presenting plausible professional output to doing the source-grounded reasoning that makes such output defensible.

\textbf{Enhancement, not replacement.} Scoping a task around the economic value and end-to-end deliverable explicitly tells a replacement story: the agent as a stand-in for the human on the job. JobBench instead scores the work whose automation augments the expert rather than substitutes for them.

Table~\ref{tab:jobbench:reporter_alignment} 
compares the design principles of JobBench against GDPVal.
For reporters, the duty experts most want offloaded is "cross-source fact checking" from the WorkBank survey \cite{shao2026futurework}
, yet GDPVal task only focuses on a single article edit over a pre-assembled source packet, while JobBench scopes the cross-year reconciliation of water-quality CSVs, EPA guidance, and surveillance data.

\begin{table*}[!t]
\centering
\scriptsize
\setlength{\tabcolsep}{2.6pt}
\renewcommand{\arraystretch}{1.10}
\caption{Case comparison between JobBench and GDPVal \citep{openai2025gdpval} on the occupations of \textit{Reporter} and \textit{Technical Sales}. }
\label{tab:jobbench:reporter_alignment}
\begin{tabularx}{\textwidth}{@{}
  >{\columncolor{black!3}\raggedright\arraybackslash}p{0.08\textwidth}
  >{\columncolor{jbblue!8}\raggedright\arraybackslash}p{0.20\textwidth}
  >{\columncolor{jbgreen!10}\raggedright\arraybackslash}p{0.135\textwidth}
  >{\columncolor{jborange!10}\raggedright\arraybackslash}p{0.25\textwidth}
  >{\columncolor{jbbrick!8}\raggedright\arraybackslash}X@{}}
\toprule
\textbf{Occupation} &
\textbf{Expert-Reported Automation Desire} &
\textbf{GDPVal} &
\textbf{JobBench} &
\textbf{Why JobBench Aligns Human Will} \\
\midrule
\textbf{Reporters} &
\textbf{\textcolor{jbblue}{Fact checking}}: check reference materials, such as books, news files, or public records, to obtain relevant facts. &
\textbf{\textcolor{jbgreen!60!black}{Article edit}}: edit a story from a source packet and return one publishable article. &
\textbf{\textcolor{jborange!90!black}{Cross-year evidence synthesis}}: cross-reference water-quality CSVs, EPA guidance, and surveillance data across years; verify threshold exceedances, identify high-risk communities, and assemble a multi-part editorial package. &
\textbf{\textcolor{jbbrick}{JobBench aligns with the real reporting burden}} by requiring cross-dataset verification before publication; \textbf{\textcolor{jbbrick}{GDPVal captures only article editing}} after the source packet has already been assembled. \\
\midrule
\textbf{Technical Sales Reps} &
\textbf{\textcolor{jbblue}{Proposal explanation}}: prepare sales presentations or proposals to explain product specifications or applications. &
\textbf{\textcolor{jbgreen!60!black}{Quote revision}}: revise a quotation from pricing and freight references. &
\textbf{\textcolor{jborange!90!black}{Bid-response package assembly}}: integrate an RFQ, site survey, internal pricing, product catalog, and competitor quote; verify certifications. &
\textbf{\textcolor{jbbrick}{JobBench aligns with the real pre-sale burden}} by requiring proposal assembly across specifications, pricing, compliance, and competitor context; \textbf{\textcolor{jbbrick}{GDPVal captures only isolated quote revision}}. \\
\bottomrule
\end{tabularx}
\renewcommand{\arraystretch}{1.0}
\end{table*}

\subsection{JobBench Overview}
\label{subsec:benchmark:overview}

\textbf{Data distribution.}
Table~\ref{tab:jobbench:stats} reports the full statistics of JobBench. It contains a 65-task main set and a 65-task easy set, covering 35 O*NET occupations spanning 10 SOC \footnote{The Standard Occupational Classification (SOC) is the U.S.~federal taxonomy that groups the labor force into occupations.} groups. Tasks are backed by 502 reference files in 17 file formats, with each 3.9 reference files on average. Most are sourced from real-world public records, including federal agency releases (e.g., CDC, EPA, EIA, Census, USDA, FRED, EEOC), state and municipal portals (city ordinances, court filings, public-health and procurement records), academic and research repositories, and open-data platforms (city open-data hubs, Kaggle, GitHub data dumps). Of the reference files in the main set, 51.7\% are from real world public records and the rest are synthesized. All reference files in the easy set come from real world.

\begin{center}
\includegraphics[width=\linewidth]{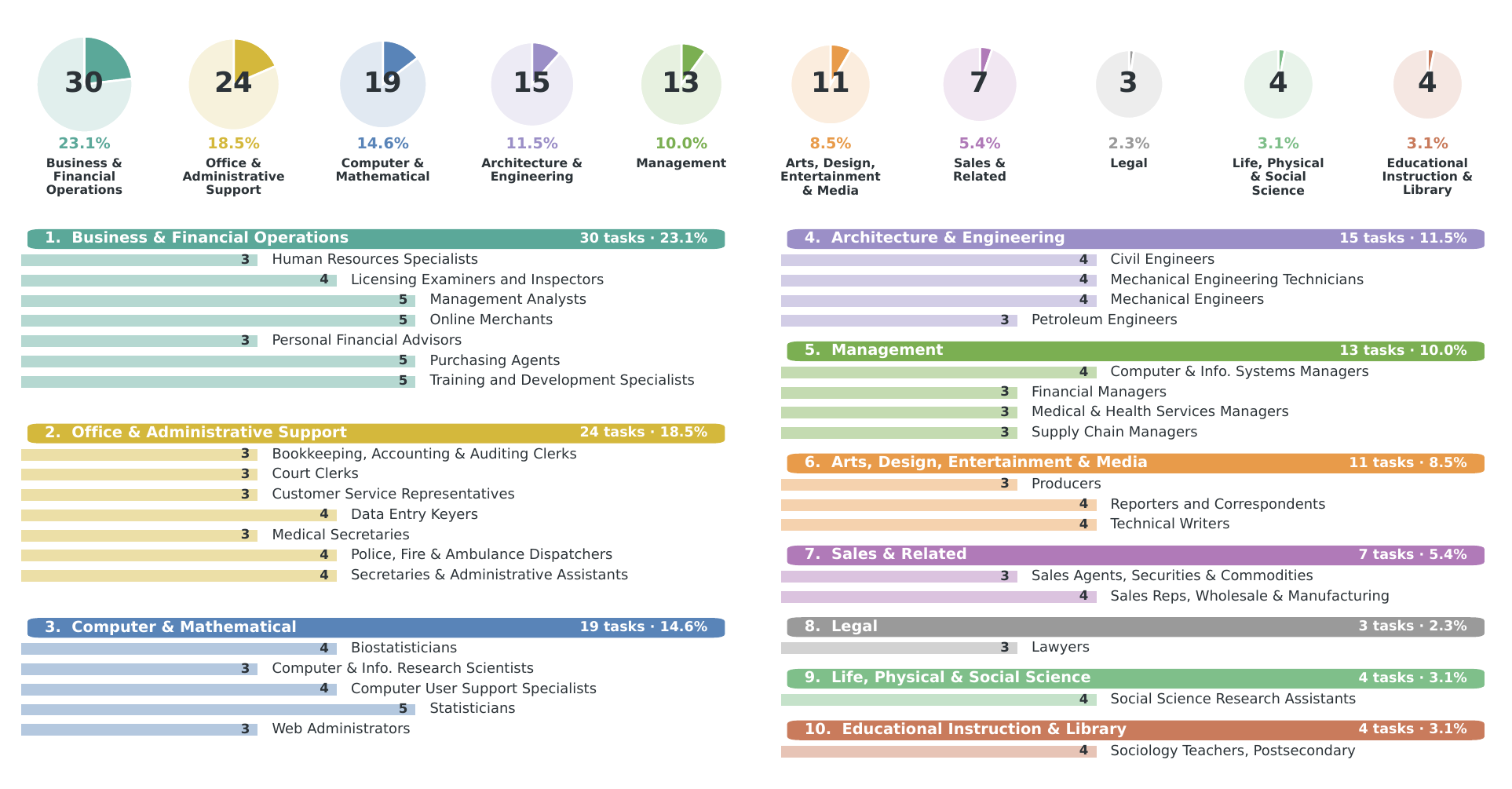}
\captionof{figure}{JobBench task distribution. Top: the 10 SOC groups covered, with each pie showing the category's share of the 130 total tasks. Bottom: the 35 occupations grouped by category, with bar length giving per-occupation task count.}
\label{fig:jobbench:task_distribution}
\end{center}

\textbf{Task specification.}
Each JobBench task is packaged as an agentic workplace bundle with:
\begin{itemize}[leftmargin=1.2em,itemsep=2pt,topsep=2pt]
    \item \textbf{Query}: A professional scenario that fixes the context and the deliverables the agent must produce.
    \item \textbf{Reference files}: A workspace of heterogeneous source documents that the agent must read across and reconcile before running the required analysis.
    \item \textbf{Binary Criterias}: Binary checks anchored to verifiable numbers, facts, and documented professional judgments.
    \item \textbf{Rubrics}: A reasoning chain whose nodes are criteria; the chain orders judgments an expert would walk to defend the underlying claim. A rubric is awarded with weighted scores only when every node passes together, so a fluent answer that quietly drops the methodology check or the threshold flagging fails the chain.
\end{itemize}
TMore task examples are shown in Appendix \ref{app:task-examples}.

\begin{wraptable}{r}{0.48\textwidth}
\vspace{-\baselineskip}
\centering
\footnotesize
\setlength{\tabcolsep}{4pt}
\renewcommand{\arraystretch}{1.15}
\caption{Key statistics of JobBench.}
\label{tab:jobbench:stats}
\begin{tabularx}{\linewidth}{@{}>{\raggedright\arraybackslash}X
                                >{\raggedleft\arraybackslash}p{0.40\linewidth}@{}}
\toprule
\textbf{Statistics} & \textbf{\#} \\
\midrule
Main / Easy set            & 65 / 65 tasks \\
Occupations / Categories   & 35 / 10 \\
\midrule
Reference files / Formats  & 502 / 17 \\
CSV / TXT / PDF / XLSX     & 171 / 25 / 153 / 81 \\
DB / DOCX / PNG / JSON     & 18 / 21 / 10 / 6 \\
Other formats              & 1--7 each \\
Files per task             & 3.9 mean; 15 max \\
\midrule
Binary criteria            & 4{,}631 \\
Criteria per task          & 35.6 mean \\
\bottomrule
\end{tabularx}
\renewcommand{\arraystretch}{1.0}
\end{wraptable}

\subsection{Benchmark Construction}
\label{subsec:benchmark:construction}

\textbf{Occupation selection.}
We target occupations that combine high automation desire with significant economic exposure. We start from Workbank, which provides crowd-sourced automation-desire scores (1-5 scale) for the O*Net work duties of each occupation \citep{shao2026futurework}, and merge its entries with OEWS~2024 total wages to quantify economic exposure \citep{bls2024oews}.
The resulting 35 occupations consist of occupations with average desire score above 3, and ranked by economic exposure. A feasibility filter then requires each retained work duty to be \emph{digitalizable}, \emph{evaluable}, and \emph{supportable} and the automation desire above 3, to form the source pool for task design. The final occupation distribution is shown in Figure~\ref{fig:jobbench:task_distribution}. 

\begin{figure}[h]
\centering
\includegraphics[width=0.95\linewidth]{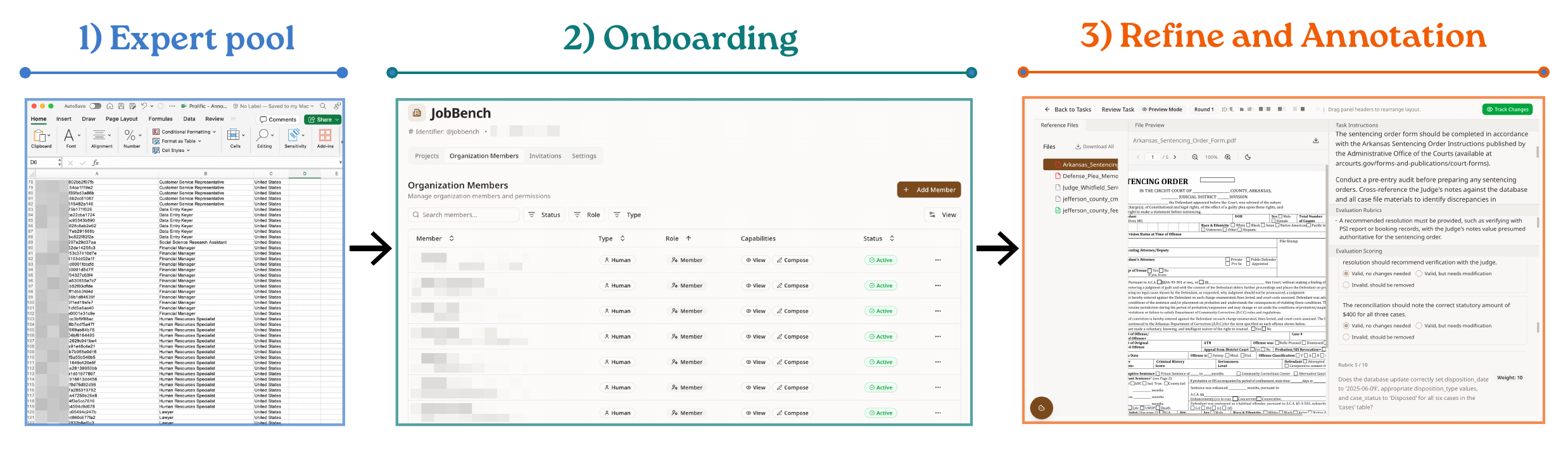}
\caption{JobBench expert onboarding: (1) the domain-expert pool from Prolific, (2) onboard experts onto the JobBench annotation platform, and (3) refine and annotation.}
\label{fig:jobbench:pipeline}
\end{figure}

\textbf{Expert pool.}
Domain experts are recruited through Prolific~\cite{palan2018prolific}, a research participant recruitment platform, and Upwork~\cite{upwork}, a freelance marketplace platform. Prolific pool covers an average of 26.5 distinct experts per occupation as indexed by participant ID. In Upwork,, we search by occupation keyword and retain candidates whose job success rate exceeds 90\%. Selected experts complete a structured onboarding flow and are then assigned jobs on our annotation platform, which integrates with AI tools used during annotation and logs the usage for review and tracking.

\textbf{Task curation.}
For each occupation, annotators draft task sketches grounded in the high-desire work duties reported in Workbank, specifying the scenario, reasoning challenges, deliverables, and the standards that should be enforced. Sketches are expanded with AI assistance into a workspace bundling reference files, self-contained task queries, and rubric chains. The Easy set tasks require no web-search evidence and pose fewer reasoning challenges than the Main set.

Every criterion in the rubric chain is required to satisfy:
\begin{itemize}[leftmargin=*,itemsep=2pt,topsep=2pt]
    \item \textbf{Self-contained}: a rubric can be judged on its own, without inheriting context outside.
    \item \textbf{Binary}: it resolves to a clean pass or fail, leaving no room for partial credit.
    \item \textbf{Objective}: it ties to a verifiable artifact or a reproducible computation.
    \item \textbf{Unambiguous}: the referent under check is named precisely enough that two graders cannot reasonably disagree on what is being evaluated. We observed that the ambiguity is the key to making different LLM judges disagree on the same answer during evaluation.
\end{itemize}

\textbf{Refine and filter.}
A candidate task passes three quality gates before entering the benchmark:
\begin{itemize}[leftmargin=*,itemsep=2pt,topsep=2pt]
    \item \textbf{Automated audit:} An audit agent checks the consistency between the task instruction and its reference files, the professional plausibility of the task itself, and the correctness of each rubric in the chain. Tasks or rubrics that fail any of these checks are dropped.
    \item \textbf{Annotator review:} Annotators polish the task instruction and prune low-quality rubrics, and only tasks that receive positive annotator feedback advance.
    \item \textbf{Solve trial:} Surviving tasks are run with different agents under multiple sampling and then judged by the rubrics. We score each task by the union of rubrics passed across all runs, and retain only tasks whose union covers more than 90\% of their own rubric set.
\end{itemize}
71\% tasks are finally passed through the 3-stage quality check pipeline. The final union pass rate across the accepted benchmark is 95.4\%, meaning that more than 95\% of all criteria were passed by at least one agent on one sample, evidence that the rubric set is achievable in practice.

%% file: sections/experiments.tex
\section{Experiments}
\label{sec:experiments}

\subsection{Experimental Setup}
\label{subsec:experiments:setup}

\begin{table*}[!tp]
\centering
\scriptsize
\setlength{\tabcolsep}{2.5pt}
\renewcommand{\arraystretch}{0.82}
\caption{JobBench-Main leaderboard across different agentic scaffolds.}
\label{tab:main_results}
\resizebox{0.7\columnwidth}{!}{
\input{analysis_outputs/jobbench_leaderboard_gdp7_min40_zero_paper.tex}
}
\renewcommand{\arraystretch}{1.0}
\end{table*}

\paragraph{Models and agentic scaffolds.}
We evaluate a representative set of recent agentic models across major proprietary
and open-source families. The evaluated models include Anthropic
Claude~\citep{anthropic2025claude4, anthropic2025claude45, anthropic2026opus46, anthropic2026opus47,
anthropic2025sonnet45, anthropic2026sonnet46, anthropic2025haiku45} (Opus-4,
Opus-4.5, Opus-4.6, Opus-4.7, Sonnet-4, Sonnet-4.5, Sonnet-4.6, and Haiku-4.5);
the OpenAI GPT-5 series~\citep{openai2025gpt5, openai2025gpt51, openai2025gpt52,
openai2026gpt54, openai2026gpt55} (GPT-5, 5.1, 5.2, 5.4, 5.5) together with its
Codex variants~\citep{openai2026codex} (GPT-5.1-Codex, 5.2-Codex, 5.3-Codex);
Google Gemini~3~\citep{google2025gemini3} (Pro and Flash);
Qwen-3.5-Plus~\citep{qwen2025qwen35}; MiniMax-M2.5~\citep{minimax2025m25};
Kimi-K2.5~\citep{moonshot2025kimik25}; and xAI
Grok-4.2-Fast~\citep{xai2025grok42}. 

We use four agentic scaffolds that span the major deployment surfaces: \texttt{Claude Code}~(v2.1.2)~\citep{anthropic2025claudecode},
\texttt{Codex CLI} (v0.125.0)~\citep{openai2025codexcli},
\texttt{OpenCode} (v1.14.18)~\citep{sst2025opencode},
and \texttt{OpenClaw} (v2026.3.8)~\citep{openclaw2025}.
Each scaffold wraps a base model with its own tool-use, planning, and file-edit policies, including shell execution, multi-file editing, sub-agent delegation, context compression, web browsing and fetching. We always choose the maximum reasoning-effort level that each model and scaffold supports by default. The agent evaluation prompt is reported in Appendix~\ref{app:prompts:agent}.

\paragraph{Task execution.}
Each task is presented as an isolated workspace, containing the local reference documents, a task instruction, and a dedicated output directory. The agent reads the instructions, operates only within this workspace, and outputs its final deliverables to the output directory. Tasks run in non-interactive headless agentic mode. Agents runners use a 60-minute wall-clock timeout per task and grant the agent file access only to the task's temporary workspace. We use each scaffold's default sampling configuration. 

\paragraph{Evaluation and judge.}
Each task is graded with rubrics by LLM-as-a-judge. The judge evaluates each rubric jointly over its binary criteria. A rubric receives its full weight $w_r$ only if every criterion passes; with the binary indicator $z_r\in\{0,1\}$, the per-task normalized score is $s_t = \sum_r w_r z_r / \sum_r w_r$. The leaderboard reports the average $\frac{1}{N}\sum_{t=1}^{N} s_t$ as the model performance. The judge prompt is shown in Appendix~\ref{app:prompts:judge}. For cost consideration, we use \texttt{x-ai/grok-4.1-fast} as the judge by default. We validate the judge results against a stronger reference model, Opus-4.5. The two judges agree on scores within 0.7\% variance across different configurations. It reduced a full-run judge API cost from \$38--\$46 with the Opus~4.5 judge to \$1.5--\$2 with Grok-4.1-Fast, an approximately $20\times$ reduction. See Section \ref{subsec:experiments:main} for more details.

\begin{figure*}[t]
\centering
\includegraphics[width=0.9\textwidth]{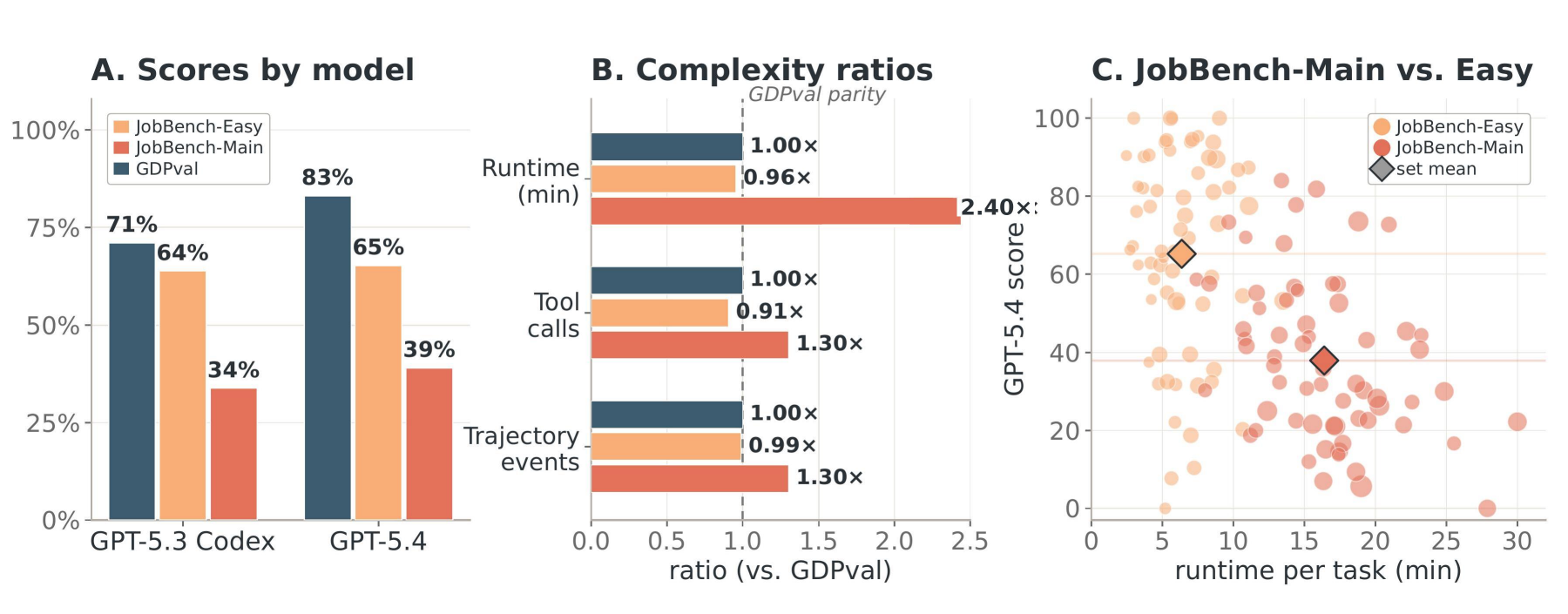}

\caption{Comparison between GDPVal and JobBench. (a) Model scores on GDPVal, JobBench-Easy, and JobBench-Main. GDPval scores approach saturation above 70 while JobBench-Main is underneath 40\% and JobBench-Easy lands in between. (b)~Runtime, tool-call, and trajectory lines complexity for GDPVal, JobBench-Easy, and JobBench-Main, normalized to GDPval~$=1.0$. (c)~For GPT-5.4, JobBench-Main tasks shift toward lower scores and longer runtimes than the easy set.}
\label{fig:jobbench:gdpval_gap}
\label{fig:jobbench:main_easy_complexity}
\end{figure*}

\subsection{Main Results}
\label{subsec:experiments:main}

Table~\ref{tab:main_results} reports the JobBench leaderboard across 36 model--scaffold configurations on the main set. We merge the four smallest occupational categories: \textit{Sales and Related}, \textit{Legal}, \textit{Life, Physical, and Social Science}, and \textit{Educational Instruction and Library}, into a single \textit{Others} column.

\paragraph{Challenging nature of JobBench.}
As shown in Table \ref{tab:main_results}, as shown in Table \ref{tab:main_results}, agents still fall short of the tasks that that experts actually want AI to perform. The strongest configuration, Claude Opus~4.7 under \texttt{Claude Code}, reaches only 45.9 on JobBench, and the next two rows---GPT-5.5 under \texttt{Codex} (42.7) and GPT-5.4 under \texttt{Codex} (38.9)---also remain below 50. Beyond the Claude and GPT families, no configuration exceeds 19 points, and the weakest, Grok-4.2-Fast, scores just 4.38. Today's agents continue to make substantial errors on the complex, professional reasoning that experts most want AI to handle.

\paragraph{GDPVal vs.\ JobBench.}
Figure~\ref{fig:jobbench:gdpval_gap} (a) shows that GDPval has approached saturation: GPT-5.3-Codex reaches 70.9 and GPT-5.4 reaches 83.0.\footnote{GDPval scores are taken from the official OpenAI report~\citep{openai2025gdpval} and correspond to the win+tie rate against industry-expert reference deliverables.} The corresponding JobBench main-set \texttt{Codex} scores, by contrast, are only 33.7 and 38.9. Figure~\ref{fig:jobbench:gdpval_gap} (b) further shows that, on JobBench-Main, GPT-5.4 under \texttt{Codex} takes $2.40\times$ the runtime of GDPval, while tool calls and trajectory events both rise to roughly $1.3\times$ GDPval levels, indicating that JobBench-Main demands substantially greater agentic work complexity.

\paragraph{Main vs. Easy}
We compare the main against the easy set with the same occupation coverage. Figure~\ref{fig:jobbench:gdpval_gap} (c) shows the shift. Scores rise by 26--31 points on the easy set, and GPT-5.4 \texttt{Codex} traces shorten from 16.4 to 6.4 minutes of runtime. Recall that the easy set by design has fewer reasoning challenges, multi-source file conflicts, deliverable files, and requires no web search during task completion -- all reference files are under the local workspace.

\begin{wrapfigure}{r}{0.42\textwidth}
\vspace{-0.8\baselineskip}
\centering
\includegraphics[width=\linewidth]{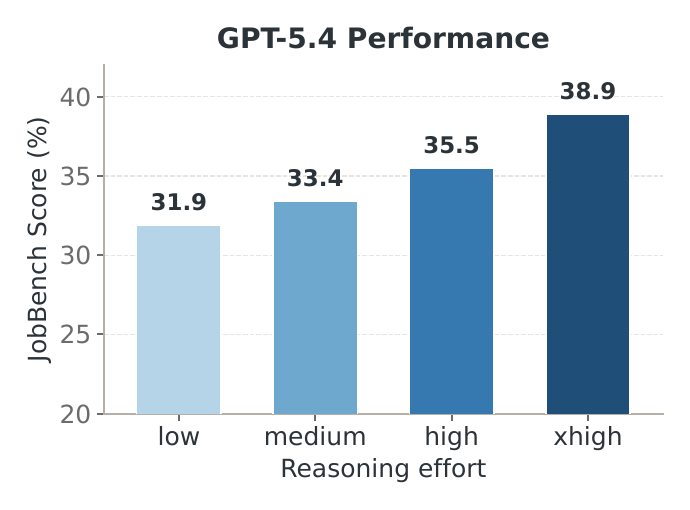}
\caption{GPT-5.4 under \texttt{Codex} on JobBench-Main as reasoning effort scales from \texttt{low} to \texttt{xhigh}.}
\label{fig:jobbench:gpt54_reasoning_effort}
\vspace{-2.0\baselineskip}
\end{wrapfigure}

\paragraph{Reasoning effort.}
Figure~\ref{fig:jobbench:gpt54_reasoning_effort} shows sweeping GPT-5.4 under \texttt{Codex} from \texttt{low} to \texttt{xhigh} reasoning effort yields a monotonic JobBench Main performance gain of $+7.0$ points.

\begin{figure*}[t]
\centering
\includegraphics[width=0.9\textwidth]{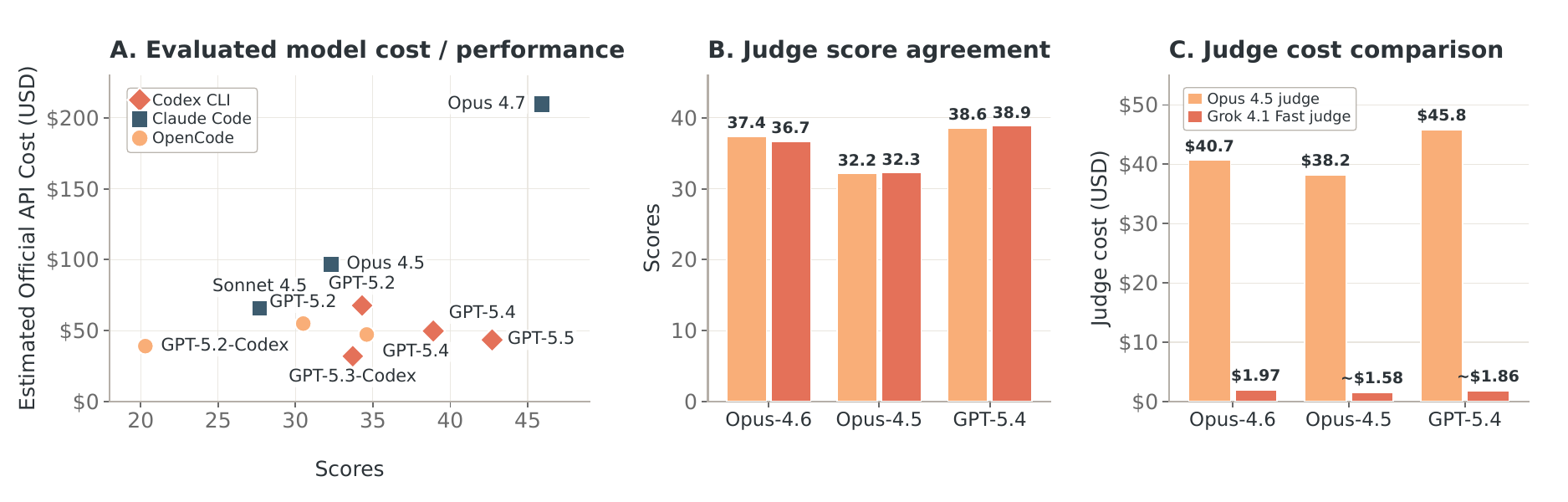}
\caption{Cost analysis for JobBench evaluation. (a)~Full-run inference cost (USD) on the main set, computed by provider list pricing. (b) Judge score agreement and (c) full-run judge cost for Claude Opus~4.5 and Grok-4.1-Fast as the judge. Grok-4.1-Fast tracks the Opus~4.5 judge within 0.1--0.7 points variance, while reducing full-run judge cost by roughly $20\times$, from ~\$40 to \$2.}
\label{fig:jobbench:diagnostics}
\end{figure*}

\paragraph{Scaffold matters as much as base model.}
Scaffold choice can shift the overall score by several points when the base model is fixed. Claude Sonnet~4.6 scores 36.9 under \texttt{Claude Code} but only 30.6 under \texttt{OpenClaw}; Claude Opus~4.5 scores 32.3 under \texttt{Claude Code} versus 29.1 under \texttt{OpenCode}; GPT-5.4 scores 38.9 under \texttt{Codex} versus 34.6 under \texttt{OpenCode}. The induced gaps are largest in \textit{Office / Admin Support}, \textit{Computer / Mathematical}, and \textit{Management}.

\paragraph{Performance vs. Cost.}
Figure~\ref{fig:jobbench:diagnostics} (a) plots overall JobBench-Main score against full-run inference cost at provider list pricing. Costs span from \$32 for GPT-5.3-Codex under \texttt{Codex CLI} to \$210 for Claude Opus-4.7 under \texttt{Claude Code}. GPT-5.5 dominates the mid-tier with 42.7 points at \$44 (cheaper \emph{and} higher-scoring in the same scaffold), and Opus-4.7 is the only configuration above 43 points but at a $\sim$$5\times$ premium over GPT-5.5.

\paragraph{Judge cost optimization.}
Figure~\ref{fig:jobbench:diagnostics} (b) and (c) compare Claude Opus-4.5 and Grok-4.1-Fast as the rubric judge under different model--scaffold configurations.  Figure~\ref{fig:jobbench:diagnostics} (b) reports that the two judges agree on judge scores within 0.7 point variance, while Figure~\ref{fig:jobbench:diagnostics} (c) shows Grok-4.1-Fast reduces full-run judge cost from \$38--\$46 (Opus~4.5) to \$1.97---an approximately $20\times$ reduction.

\subsection{Occupational Analysis}
\label{subsubsec:experiments:occupation}

\begin{figure*}[t]
\centering
\includegraphics[width=\textwidth]{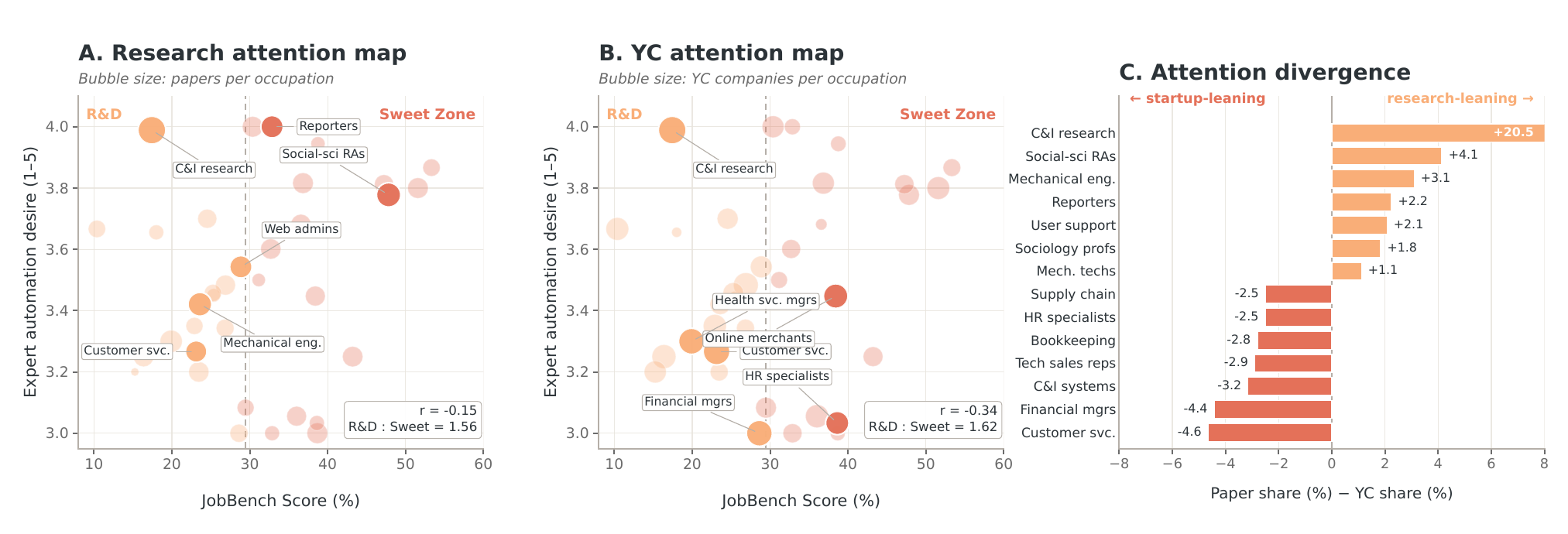}
\caption{Occupation analysis for research and startup attention over JobBench. Each occupation is placed by JobBench-Main scores on the x-axis (mean across all Claude and GPT family models) and by the WorkBank expert automation-desire score on the y-axis; dashed median splits define the high capability \emph{Sweet Zone} and low capability \emph{Research \& Develop (R\&D)} quadrants. (a)~Bubble size encodes the count of agentic-LLM arXiv papers (April 2025--March 2026; 2{,}932 unique papers tagged by GPT-5.4-mini against the 35 high-desire occupations). (b)~Bubble size encodes the per-occupation count of AI-related YC companies (batches 2006--W26; 2{,}089 unique companies). Attention correlates \emph{negatively} with capability ($r=-0.15$ for papers, $r=-0.34$ for YC), and the R\&D-to-Sweet-Zone attention ratio exceeds one ($1.56$ and $1.62$); external attention therefore concentrates more on the R\&D-quadrant than the Sweet-Zone, where agents already handle well. (c)~Attention divergence: Research leans toward knowledge-heavy occupations, while startups instead concentrate on more economically-significant areas.}
\label{fig:jobbench:strategy_dashboard}
\end{figure*}

\paragraph{Setup.}
We further study how the research and startup communities allocate attention across high-automation-desire occupations. We tag 3,516 agentic-related LLM arXiv paper abstracts (April 2025--March 2026) and 2,283 AI-related YC company descriptions (batches 2006--W26) with GPT-5.4-mini, deciding for each whether it falls within the 35 high-desire occupations and corresponding work duties covered by JobBench. This yields 2,932 papers and 2,089 startups labeled with at least one matching occupation; a single paper or company may map to multiple occupations. For each occupation, we plot its mean JobBench-Main score (averaged across all Claude and GPT family models) against its WORKBank automation-desire score in Figure~\ref{fig:jobbench:strategy_dashboard}, with the per-occupation paper/ startup count encoded as bubble size. The median split of capability against desire defines the high-capability \emph{Sweet Zone} and low-capability \emph{Research \& Develop (R\&D)} quadrants. We compute the Pearson correlation $r$ between attention (paper/startup per occupation) and JobBench score.

\paragraph{Both research papers and startups concentrate more on the R\&D quadrant.}
Figure~\ref{fig:jobbench:strategy_dashboard} (a) and (b) show that attention correlates \emph{negatively} with model capability ($r=-0.15$ for papers, $r=-0.34$ for YC), and the R\&D-to-Sweet-Zone attention ratio both exceeds one ($1.56$ for papers, $1.62$ for YC). In other words, R\&D-quadrant occupations attract more attention than the Sweet-Zone ones where agents already handle well. Figure~\ref{fig:jobbench:strategy_dashboard} (c) decomposes this attention into a per-occupation gap, isolating where the two communities focus differently on the same areas. Research piles disproportionately onto \texttt{computer\_and\_information\_research\_scientists} ($+20.5$~pp gap), likely reflecting the boom in LLM-related research these years, and also leans toward knowledge-heavy occupations such as \texttt{social\_science\_research\_assistants}. Startups instead concentrate on \texttt{customer\_service\_representatives} ($+4.6$), and \texttt{financial\_managers} ($+4.4$), spreading attention toward more economically lucrative occupations.

%% file: analysis_outputs/jobbench_leaderboard_gdp7_min40_zero_paper.tex
\begin{tabular}{@{}lrrrrrrrr@{}}
\toprule
\rowcolor{black!4}\cell{Model}{~} & \cell{Overall}{~} & \cell{Bus.}{Fin.} & \cell{Admin}{~} & \cell{Comp.}{Math.} & \cell{Arch.}{Eng.} & \cell{Mgmt.}{~} & \cell{Arts}{~} & \cell{Others}{~} \\
\midrule
\rowcolor{black!6}\multicolumn{9}{@{}l}{\textbf{\leaderlogo{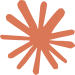} Claude Code}} \\
\leaderlogo{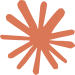} Opus-4.7 & \textbf{45.9} & \underline{46.1} & \textbf{47.8} & \underline{39.2} & \textbf{46.6} & \textbf{38.8} & \textbf{64.2} & \underline{46.3} \\
\leaderlogo{assets/logos/anthropic.png} Sonnet-4.6 & 36.9 & 36.7 & 38.4 & 31.9 & 41.1 & 30.7 & \underline{54.9} & 33.1 \\
\leaderlogo{assets/logos/anthropic.png} Opus-4.6 & 36.7 & 33.7 & \underline{47.3} & 38 & 38.8 & 29.2 & 42.8 & 30 \\
\leaderlogo{assets/logos/anthropic.png} Opus-4.5 & 32.3 & 30.1 & 33.3 & 32.7 & 36.7 & 16.7 & 42.3 & 38.1 \\
\leaderlogo{assets/logos/anthropic.png} Sonnet-4.5 & 27.7 & 23.9 & 19 & 35.9 & 30.9 & 18.1 & 46 & 29.6 \\
\leaderlogo{assets/logos/anthropic.png} Opus-4 & 21.9 & 27.8 & 16 & 21.4 & 14.7 & 15.2 & 35.6 & 24.5 \\
\leaderlogo{assets/logos/anthropic.png} Sonnet-4 & 18.4 & 16.5 & 25.3 & 21.1 & 11.2 & 11.7 & 33.6 & 15.2 \\
\leaderlogo{assets/logos/anthropic.png} Haiku-4.5 & 16 & 11.7 & 17.8 & 19.8 & 15.4 & 9.13 & 32.2 & 14.4 \\
\midrule
\rowcolor{black!6}\multicolumn{9}{@{}l}{\textbf{\leaderlogo{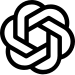} Codex CLI}} \\
\leaderlogo{assets/logos/openai_logo.png} GPT-5.5 & \underline{42.7} & \textbf{47.7} & 42.6 & \textbf{39.4} & 40.9 & 30.5 & 50.2 & \textbf{46.5} \\
\leaderlogo{assets/logos/openai_logo.png} GPT-5.4 & 38.9 & 38.2 & 44.8 & 34.6 & 35.9 & 23.5 & 51.8 & 46.2 \\
\leaderlogo{assets/logos/openai_logo.png} GPT-5.2 & 34.3 & 30.1 & 39.8 & 31.6 & 31.4 & 26.7 & 45.7 & 40.1 \\
\leaderlogo{assets/logos/openai_logo.png} GPT-5.3-Codex & 33.7 & 31.4 & 40.3 & 35.1 & 22.9 & 20.1 & 48.7 & 40.8 \\
\leaderlogo{assets/logos/openai_logo.png} GPT-5.1-Codex & 26.2 & 21.2 & 27.6 & 29.4 & 28.1 & 11.9 & 44.2 & 29.4 \\
\leaderlogo{assets/logos/openai_logo.png} GPT-5.2-Codex & 26 & 19 & 28.8 & 24.1 & 25.3 & 14.9 & 37.9 & 38.6 \\
\midrule
\rowcolor{black!6}\multicolumn{9}{@{}l}{\textbf{\leaderlogo{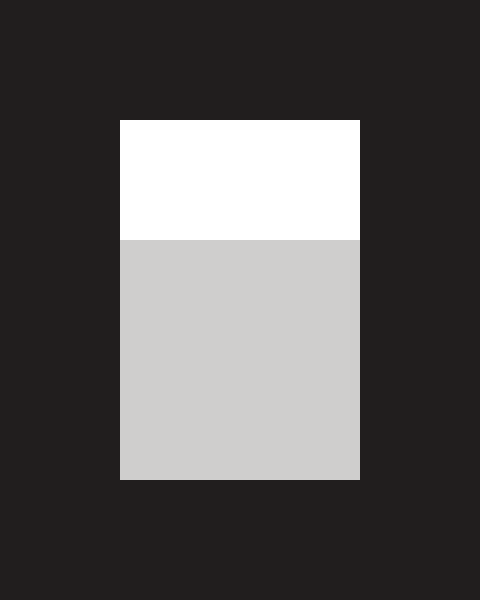} OpenCode}} \\
\leaderlogo{assets/logos/openai_logo.png} GPT-5.4 & 34.6 & 28.9 & 33.9 & 36.2 & \underline{41.6} & 14.1 & 49.5 & 45.3 \\
\leaderlogo{assets/logos/openai_logo.png} GPT-5.2 & 30.5 & 34.7 & 23.7 & 25.7 & 29.9 & 16.7 & 53 & 36.7 \\
\leaderlogo{assets/logos/anthropic.png} Opus-4.5 & 29.1 & 26.4 & 25.1 & 26.8 & 30.4 & 19.1 & 42 & 39.6 \\
\leaderlogo{assets/logos/anthropic.png} Sonnet-4.5 & 22 & 20.3 & 22.3 & 27.4 & 22.2 & 13.2 & 38.8 & 17.2 \\
\leaderlogo{assets/logos/openai_logo.png} GPT-5.2-Codex & 20.3 & 11.3 & 16.3 & 18.2 & 24.9 & 15.6 & 33.1 & 32.4 \\
\leaderlogo{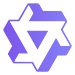} Qwen-3.5-Plus & 18.5 & 15.1 & 20.9 & 22.8 & 15.8 & 10.3 & 34.4 & 17.5 \\
\leaderlogo{assets/logos/openai_logo.png} GPT-5.1 & 16.3 & 17.2 & 14.9 & 17.5 & 9.03 & 12.1 & 36.7 & 14.3 \\
\leaderlogo{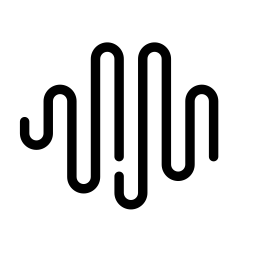} MiniMax-M2.5 & 14.8 & 12.2 & 15.4 & 18.1 & 14.2 & 6.74 & 22.5 & 17.6 \\
\leaderlogo{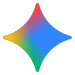} Gemini-3-Pro & 11.4 & 11.8 & 10.7 & 12 & 12.9 & 8.71 & 11.3 & 11.7 \\
\leaderlogo{assets/logos/gemini.png} Gemini-3-Flash & 11.4 & 11.2 & 13.9 & 10.1 & 14.7 & 10.8 & 7.45 & 10.2 \\
\leaderlogo{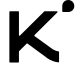} Kimi-K2.5 & 8.73 & 6.68 & 8.16 & 10.9 & 8.11 & 3.06 & 17.1 & 10.8 \\
\leaderlogo{assets/logos/openai_logo.png} GPT-5 & 8.53 & 6.8 & 9.62 & 11.5 & 4.01 & 0.87 & 15.4 & 13.2 \\
\leaderlogo{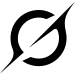} Grok-4.2-Fast & 4.38 & 2.6 & 3.33 & 5.06 & 3.56 & 3.02 & 10.4 & 5.97 \\
\midrule
\rowcolor{black!6}\multicolumn{9}{@{}l}{\textbf{\leaderlogo{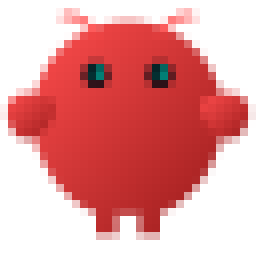} OpenClaw}} \\
\leaderlogo{assets/logos/anthropic.png} Opus-4.6 & 36.6 & 34.1 & 43 & 25 & 37.6 & \underline{33} & 44.8 & 43.2 \\
\leaderlogo{assets/logos/openai_logo.png} GPT-5.4 & 33.1 & 26.6 & 35.7 & 32.4 & 40 & 30.2 & 35.9 & 36 \\
\leaderlogo{assets/logos/anthropic.png} Sonnet-4.6 & 30.6 & 30.9 & 33 & 20 & 30.2 & 23.8 & 48.7 & 35.4 \\
\leaderlogo{assets/logos/anthropic.png} Opus-4.5 & 28 & 23.3 & 29.9 & 21.5 & 35.4 & 13 & 49.2 & 34.6 \\
\leaderlogo{assets/logos/openai_logo.png} GPT-5.2 & 27.6 & 25.6 & 30.7 & 15.7 & 19.1 & 29.4 & 43 & 36.6 \\
\leaderlogo{assets/logos/anthropic.png} Sonnet-4.5 & 24.5 & 16.4 & 28.6 & 29.3 & 22 & 20 & 36 & 27.1 \\
\leaderlogo{assets/logos/openai_logo.png} GPT-5.3-Codex & 17.8 & 14.2 & 18.5 & 21.3 & 17.5 & 9.64 & 20 & 24 \\
\leaderlogo{assets/logos/anthropic.png} Haiku-4.5 & 17.4 & 16.4 & 16.2 & 15 & 19.5 & 8.27 & 36.6 & 18.4 \\
\leaderlogo{assets/logos/openai_logo.png} GPT-5.2-Codex & 16.5 & 16 & 18.6 & 12.8 & 13.7 & 7.68 & 23.6 & 24.7 \\
\bottomrule
\end{tabular}

%% file: sections/conclusion.tex
\section{Conclusion}
\label{sec:conclusion}

In this paper, we introduce JobBench, a benchmark that aligns agentic evaluation with human will instead of only focusing on economic values. Its 130 tasks span 35 occupations, each built from a Workbank-elicited delegation preference, packaged as a workspace of heterogeneous reference files, and graded by chained rubrics whose 4,631 binary criteria award credit only when every step in the chain holds together. 
Across 36 models, the strongest setup, Claude Opus~4.7 under \texttt{Claude Code}, reaches 45.9, and outside the Claude and GPT families, no models exceed 20\%. JobBench is a leaderboard for tracking agent capability on work duties whose automation would most directly enhance the workers' satisfaction and productivity. We hope it shifts the community's labour-market target from replacement to enhancement, building agents that do what humans actually want delegated.

\section*{Acknowledgment}

This work is partially supported by the National Science Foundation (NSF) AI Institute for Agent-based Cyber Threat Intelligence and Operation (ACTION) under grant IIS 2229876, the Office of Naval Research (ONR) under grant N0014-23-1-2386, and the Air Force Office of Scientific Research (AFOSR) under grant FA9550-23-1-0208.

This work is supported in part by funds provided by the National Science Foundation, Department of Homeland Security, and IBM. 
Any opinions, findings, and conclusions or recommendations expressed in this material are those of the author(s) and do not necessarily reflect the views of the NSF or its federal agency and industry partners.

%% file: sections/appendix_toc.tex
\begingroup
\setlength{\parindent}{0pt}

\newcommand{\tocSec}[3]{\noindent\hbox to \textwidth{\makebox[2.0em][l]{\textbf{\large\textcolor{jbblue}{#1}}}\textbf{\large #2}\,\xleaders\hbox{\hskip 0.45em.\hskip 0.45em}\hfill\,\textbf{\pageref{#3}}}\par\addvspace{0.55em}}

\newcommand{\tocSecHead}[2]{\noindent\hbox to \textwidth{\makebox[2.0em][l]{\textbf{\large\textcolor{jbblue}{#1}}}\textbf{\large #2}\hfill}\par\addvspace{0.30em}}

\newcommand{\tocSub}[3]{\noindent\hbox to \textwidth{\hspace*{2.0em}\makebox[2.6em][l]{\textcolor{jbblue!75!black}{#1}}{#2}\,\xleaders\hbox{\hskip 0.40em\textcolor{black!55}{.}\hskip 0.40em}\hfill\,\pageref{#3}}\par\addvspace{0.30em}}

\tocSec{A}{Related Work}{sec:related_work}
\tocSec{B}{Limitations, Ethics, Broader Impact, and LLM Usage}{app:responsible_use}
\tocSec{C}{Leaderboard}{app:main-leaderboard}
\tocSec{D}{Task Split by Occupation}{app:task-split}

\tocSecHead{E}{Representative Task Examples}
\tocSub{E.1}{Reporters and Correspondents}{app:task-example:reporters}
\tocSub{E.2}{Data Entry Keyers}{app:task-example:data-entry}
\tocSub{E.3}{Lawyers}{app:task-example:lawyers}
\tocSub{E.4}{Web Administrators}{app:task-example:web-admin}
\addvspace{0.20em}

\tocSecHead{F}{Prompt Templates and Evaluation Interface}
\tocSub{F.1}{Agent Evaluation Prompt}{app:prompts:agent}
\tocSub{F.2}{Judge Prompt}{app:prompts:judge}
\addvspace{0.20em}

\tocSec{G}{Supplementary Analysis on JobBench's Position in AI Labour Market}{app:supplementary}

\endgroup

%% file: sections/related_work.tex
\section{Related Work}
\label{sec:related_work}

\paragraph{Frontier knowledge and reasoning benchmarks.}
Many benchmarks score whether a model reaches the correct answer on
harder exam-style or code-style prompts. MMLU \citep{hendrycks2021mmlu}
established a 57-subject baseline, GPQA \citep{rein2024gpqa} pushes to
graduate-level science designed to resist Web search, and MMLU-Pro
\citep{wang2024mmlupro} expands the answer space to reward multi-step
reasoning. MMMU \citep{yue2024mmmu} extends expert-level evaluation to
multimodal inputs. 

\paragraph{Agentic benchmarks in interactive environments.}
A parallel line of work targets agents that act, rather than answer.
SWE-bench \citep{jimenez2024swebench} measures whether agents resolve
real GitHub issues with a working patch, WebArena and WorkArena
\citep{zhou2024webarena,drouin2024workarena} instantiate enterprise web
stacks, and OSWorld \citep{xie2024osworld} extends the setting to full
desktop environments. GAIA \citep{mialon2024gaia}, AgentBench \citep{liu2024agentbench}, WebShop \citep{yao2022webshop}, and $\tau$-bench
\citep{yao2024taubench} cover assistant tool use, multi-environment
generally, grounded shopping, and tool-agent-user interaction. These
benchmarks score whether an agent can \emph{reach} a target state in an
environment

\paragraph{Workplace and occupational benchmarks.}
Recent benchmarks target the work that experts actually perform. GDPVal \citep{openai2025gdpval} frames AI progress in terms of economically valuable tasks judged by expert graders, PRBench \citep{akyurek2025prbench} supplies expert-authored rubrics for legal and financial reasoning, and the Remote Labor Index \citep{mazeika2025rli} scores end-to-end remote-work projects and
surfaces deliverable-level failure clusters. 
These benchmarks are scoped primarily by economic values, telling a replacement story. JobBench instead evaluates AI agents on the workflows that experts identify as high-priority for delegation, empowering humans based on their needs instead of replacing them with GDP value.

\paragraph{Labor-market exposure.}
Recent work measures how AI maps onto the U.S.\ workforce.
\citet{eloundou2023gpts} estimate task-level occupational exposure to
LLMs, \citet{brynjolfsson2023genai} measure productivity gains from a
customer-support deployment, and \citet{chen2024displacement} study
early displacement and complementarity effects in the labor market.
Closer to task design, Workbank \citep{shao2026futurework} surveys
over 1{,}500 U.S.\ workers on the O*NET task taxonomy and labels
each task with a reported delegation preference and a desired human-AI
collaboration mode. However, it only covers worker surveys for the specific work duties. JobBench is built on top of these signals and designs them into benchmark task packages. 

%% file: sections/discussion.tex
\section{Limitations, Ethics, Broader Impact, and LLM Usage}
\label{app:responsible_use}

\subsection*{Limitations}
This study is limited to U.S.-centered, digital, document-heavy professional tasks across 35 selected O*NET occupations. It does not represent all occupations, non-U.S. labor markets, non-English workplaces, physical work, real-time collaboration, or long-term organizational workflows. The dataset is designed for benchmark evaluation, not deployment validation. It is strictly not recommended for deciding whether an AI system can replace workers, make professional judgments, or operate without human oversight in legal, medical, financial, engineering, public-sector, or other high-stakes settings.

\subsection*{Ethics Statement}
This study involved human participants providing data annotation through an online annotation platform. The University Human Subjects Division (HSD) reviewed the protocol and determined that the research qualifies as exempt human subjects research (Category 3; minimal-risk behavioral research) with approved IRB under U.S. federal regulations. Participation was voluntary, and participants could stop at any time. No sensitive personal information was collected. Only aggregated benchmark results are publicly released, and no identifiable participant data are included in any publications or datasets. The exempt determination letter is on file with the authors. All annotator participating were informed of and consented to the use of their work for research. We are committed to crediting and fairly compensating all human annotators in accordance with occupational wage standards.

This study is not intended to include direct personal identifiers of private individuals. Names and person-specific details of private individuals appearing in task text or reference files are fictional, redacted, or replaced with synthetic placeholders. Real reference files, where included, are derived from public records, official materials, or public reporting. These files may describe public activities of public officials acting in their official capacity or appearing in public contexts. The dataset may still contain geography, language, socio-economic status, experience or seniority, occupational context,  sourced from public records with masked or redacted personally identifiable information. Public references are included only as contextual source material and should not be used for impersonation, profiling, harassment, surveillance, or making decisions.

\subsection*{Broader Impact}
Positive impact: This study supports the evaluation of AI agents on work that professionals report wanting help with, encouraging augmentation rather than replacement, and highlighting where current systems still fail on realistic workplace tasks. 

Risks: the dataset could be misused as evidence that agents are ready to replace workers, or as a proxy for safe deployment in legal, financial, engineering, or public-sector settings. Because it is U.S.-centric and document-heavy, it may underrepresent non-U.S., non-English, physical-labor, and low-resource work contexts. 

Mitigations: This study is framed as an evaluation benchmark only, with explicit limitations against deployment certification or worker replacement claims. The release should include usage terms, data provenance notes, and guidance requiring expert validation and human oversight for any high-stakes use.

\subsection*{LLM Usage}
We used LLMs to support the presentation of this manuscript, including assistance with writing, editing, and improving the clarity of presentation.

%% file: sections/appendix_leaderboard.tex
\section{Leaderboard}
\label{app:main-leaderboard}

We present the leaderboard for the JobBench main set. Figure~\ref{fig:jobbench:leaderboard_vertical} ranks 15 models on the main set.

\begin{center}
\centering
\includegraphics[width=\textwidth]{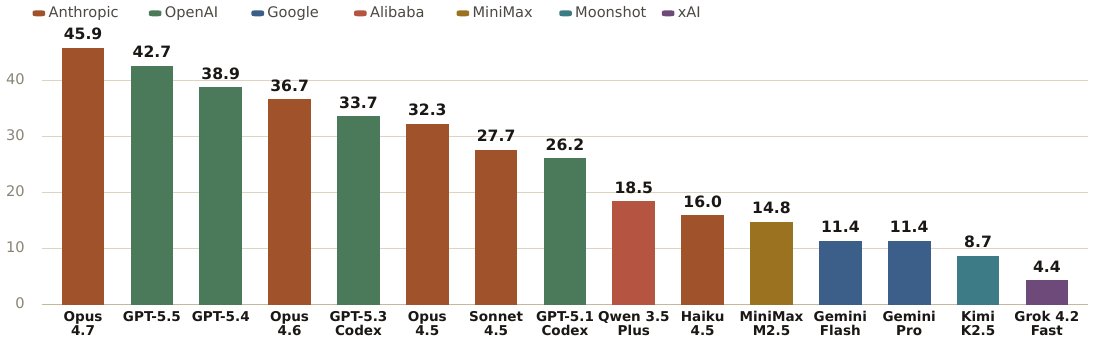}
\captionof{figure}{Leaderboard on the JobBench main set (Claude Code for Anthropic models, Codex for OpenAI models, OpenCode for the remaining models).}
\label{fig:jobbench:leaderboard_vertical}
\end{center}

%% file: analysis_outputs/jobbench_task_split_appendix.tex
\section{Task Split by Occupation}
\label{app:task-split}

In this appendix, we report the per-occupation task split. Table~\ref{tab:jobbench:task_split} reports the number of tasks each occupation contributes to the 65-task main set and the 65-task easy set.

\begin{center}
\scriptsize
\setlength{\tabcolsep}{4pt}
\renewcommand{\arraystretch}{1.05}
\captionof{table}{Per-occupation task counts on the JobBench.}
\label{tab:jobbench:task_split}
\begin{tabularx}{\textwidth}{@{}>{\raggedright\arraybackslash}X>{\raggedleft\arraybackslash}p{0.10\textwidth}>{\raggedleft\arraybackslash}p{0.10\textwidth}>{\raggedleft\arraybackslash}p{0.10\textwidth}@{}}
\toprule
\textbf{Occupation} & \textbf{Main} & \textbf{Easy} & \textbf{Total} \\
\midrule
\rowcolor{jbblue!7}\multicolumn{4}{@{}l@{}}{\textbf{Business and Financial Operations} (\textit{14 Main / 16 Easy / 30 Total})} \\
Human Resources Specialists & 1 & 2 & 3 \\
Licensing Examiners and Inspectors & 1 & 3 & 4 \\
Management Analysts & 3 & 2 & 5 \\
Online Merchants & 2 & 3 & 5 \\
Personal Financial Advisors & 1 & 2 & 3 \\
Purchasing Agents, Except Wholesale, Retail, and Farm Products & 3 & 2 & 5 \\
Training and Development Specialists & 3 & 2 & 5 \\
\addlinespace[2pt]
\rowcolor{jbblue!7}\multicolumn{4}{@{}l@{}}{\textbf{Office and Administrative Support} (\textit{10 Main / 14 Easy / 24 Total})} \\
Bookkeeping, Accounting, and Auditing Clerks & 2 & 1 & 3 \\
Court Clerks & 1 & 2 & 3 \\
Customer Service Representatives & 1 & 2 & 3 \\
Data Entry Keyers & 2 & 2 & 4 \\
Medical Secretaries & 1 & 2 & 3 \\
Police, Fire, and Ambulance Dispatchers & 1 & 3 & 4 \\
Secretaries and Administrative Assistants, Except Legal, Medical, and Executive & 2 & 2 & 4 \\
\addlinespace[2pt]
\rowcolor{jbblue!7}\multicolumn{4}{@{}l@{}}{\textbf{Computer and Mathematical} (\textit{10 Main / 9 Easy / 19 Total})} \\
Biostatisticians & 2 & 2 & 4 \\
Computer and Information Research Scientists & 2 & 1 & 3 \\
Computer User Support Specialists & 2 & 2 & 4 \\
Statisticians & 3 & 2 & 5 \\
Web Administrators & 1 & 2 & 3 \\
\addlinespace[2pt]
\rowcolor{jbblue!7}\multicolumn{4}{@{}l@{}}{\textbf{Architecture and Engineering} (\textit{8 Main / 7 Easy / 15 Total})} \\
Civil Engineers & 3 & 1 & 4 \\
Mechanical Engineering Technicians & 3 & 1 & 4 \\
Mechanical Engineers & 1 & 3 & 4 \\
Petroleum Engineers & 1 & 2 & 3 \\
\addlinespace[2pt]
\rowcolor{jbblue!7}\multicolumn{4}{@{}l@{}}{\textbf{Management} (\textit{8 Main / 5 Easy / 13 Total})} \\
Computer and Information Systems Managers & 2 & 2 & 4 \\
Financial managers, branch or department & 2 & 1 & 3 \\
Medical and Health Services Managers & 2 & 1 & 3 \\
Supply Chain Managers & 2 & 1 & 3 \\
\addlinespace[2pt]
\rowcolor{jbblue!7}\multicolumn{4}{@{}l@{}}{\textbf{Arts, Design, Entertainment, Sports, and Media} (\textit{5 Main / 6 Easy / 11 Total})} \\
Producers & 1 & 2 & 3 \\
Reporters and Correspondents & 1 & 3 & 4 \\
Technical Writers & 3 & 1 & 4 \\
\addlinespace[2pt]
\rowcolor{jbblue!7}\multicolumn{4}{@{}l@{}}{\textbf{Sales and Related} (\textit{3 Main / 4 Easy / 7 Total})} \\
Sales Agents, Securities and Commodities & 1 & 2 & 3 \\
Sales representatives, wholesale and manufacturing, technical and scientific products & 2 & 2 & 4 \\
\addlinespace[2pt]
\rowcolor{jbblue!7}\multicolumn{4}{@{}l@{}}{\textbf{Legal} (\textit{1 Main / 2 Easy / 3 Total})} \\
Lawyers & 1 & 2 & 3 \\
\addlinespace[2pt]
\rowcolor{jbblue!7}\multicolumn{4}{@{}l@{}}{\textbf{Life, Physical, and Social Science} (\textit{3 Main / 1 Easy / 4 Total})} \\
Social Science Research Assistants & 3 & 1 & 4 \\
\addlinespace[2pt]
\rowcolor{jbblue!7}\multicolumn{4}{@{}l@{}}{\textbf{Educational Instruction and Library} (\textit{3 Main / 1 Easy / 4 Total})} \\
Sociology Teachers, Postsecondary & 3 & 1 & 4 \\
\addlinespace[2pt]
\midrule
\textbf{All categories (35 occupations)} & \textbf{65} & \textbf{65} & \textbf{130} \\
\bottomrule
\end{tabularx}
\renewcommand{\arraystretch}{1.0}
\end{center}

%% file: sections/appendix_task_examples.tex
\section{Representative Task Examples}
\label{app:task-examples}

In this appendix, we show four representative JobBench tasks drawn from the main-set: one Reporters task (Section~\ref{app:task-example:reporters}), one Data Entry Keyers task (Section~\ref{app:task-example:data-entry}), one Lawyers task (Section~\ref{app:task-example:lawyers}), and one Web Administrators task (Section~\ref{app:task-example:web-admin}). For each task, we present three views: a one-page schematic (assignment brief, evidence files, reasoning burden, required deliverables, and rubric pressure points), the verbatim agent instruction, and a reviewer-facing parse of the task rubrics and the task card.
\subsection{Reporters: Lead Exposure Editorial Package}
\label{app:task-example:reporters}

\begin{center}
\centering
\resizebox{\textwidth}{!}{\begin{tikzpicture}[font=\sffamily]
  \draw[rounded corners=7pt, line width=1.2pt, draw=domainmedia!90!black, fill=domainmedia!8] (0,0) rectangle (16,8.6);
  \fill[domainmedia!22] (0,7.65) rectangle (16,8.6);
  \node[anchor=west, font=\Large\bfseries, text=domainmedia!90!black] at (0.35,8.13) {Reporters and Correspondents -- Task 1};
  \node[anchor=east, font=\small\bfseries, text=domainmedia!80!black] at (15.65,8.13) {Pre-publication editorial package};

  \draw[rounded corners=4pt, draw=jbblue!80!black, fill=white, line width=0.7pt] (0.45,4.45) rectangle (5.25,7.35);
  \fill[jbblue!12] (0.45,6.88) rectangle (5.25,7.35);
  \node[anchor=west, font=\bfseries\small, text=jbblue!85!black] at (0.68,7.12) {Assignment Brief};
  \node[anchor=north west, align=left, text width=4.35cm, font=\scriptsize] at (0.70,6.72) {Act as an investigative reporter at a Connecticut regional outlet. Build a front-page package on lead in drinking water and childhood health, following the editor's assignment email and checking interview claims against public data.};

  \draw[rounded corners=4pt, draw=jbgreen!75!black, fill=white, line width=0.7pt] (0.45,0.55) rectangle (5.25,4.12);
  \fill[jbgreen!13] (0.45,3.65) rectangle (5.25,4.12);
  \node[anchor=west, font=\bfseries\small, text=jbgreen!70!black] at (0.68,3.89) {Evidence Files};
  \node[anchor=north west, align=left, text width=4.35cm, font=\scriptsize] at (0.70,3.45) {
    Editor assignment email\\
    Hartford-area water lead samples CSV\\
    CT 2024 lead surveillance report\\
    NRDC Newark drinking-water factsheet\\
    Dr. Martinez interview notes
  };

  \draw[rounded corners=4pt, draw=jborange!90!black, fill=white, line width=0.7pt] (5.65,1.05) rectangle (10.35,7.15);
  \fill[jborange!15] (5.65,6.68) rectangle (10.35,7.15);
  \node[anchor=west, font=\bfseries\small, text=jborange!80!black] at (5.88,6.92) {Reasoning Burden};
  \node[anchor=north west, align=left, text width=4.15cm, font=\scriptsize] at (5.90,6.47) {
    \textbf{Cross-source reconciliation}\\[-0.2em]
    Map FOIA water samples to city-level surveillance claims.\\[0.22em]
    \textbf{Quantitative validation}\\[-0.2em]
    Compute 2020--2024 lead-level trends and CT-vs-national blood-lead comparisons.\\[0.22em]
    \textbf{Editorial grounding}\\[-0.2em]
    Pair quotes with verified data and avoid unsupported water-versus-paint claims.\\[0.22em]
    \textbf{Regulatory precision}\\[-0.2em]
    Use the EPA LCRI action-level timeline correctly.
  };

  \draw[rounded corners=4pt, draw=badgeopenai!75!black, fill=white, line width=0.7pt] (10.75,4.30) rectangle (15.55,7.15);
  \fill[badgeopenai!13] (10.75,6.68) rectangle (15.55,7.15);
  \node[anchor=west, font=\bfseries\small, text=badgeopenai!80!black] at (10.98,6.92) {Required Deliverables};
  \node[anchor=north west, align=left, text width=4.25cm, font=\scriptsize] at (11.00,6.47) {
    \texttt{pitch\_memo.docx}: thesis and evidentiary backbone.\\[0.28em]
    \texttt{data\_analysis.xlsx}: water trends, CT vs national, and city cross-reference sheets.\\[0.28em]
    \texttt{source\_log.csv}: 15+ verification entries across all sources.
  };

  \draw[rounded corners=4pt, draw=jbbrick!90!black, fill=white, line width=0.7pt] (10.75,0.85) rectangle (15.55,3.90);
  \fill[jbbrick!13] (10.75,3.43) rectangle (15.55,3.90);
  \node[anchor=west, font=\bfseries\small, text=jbbrick!90!black] at (10.98,3.67) {Rubric Pressure Points};
  \node[anchor=north west, align=left, text width=4.25cm, font=\scriptsize] at (11.00,3.22) {High-risk community mapping; lead-level percentage changes; CT surveillance extraction; Newark numerical parallel; interview claim verification; 15-entry source log coverage; workbook sheet completeness.};

  \draw[->, line width=0.8pt, draw=domainmedia!75!black] (5.25,5.90) -- (5.65,5.90);
  \draw[->, line width=0.8pt, draw=domainmedia!75!black] (5.25,2.20) -- (5.65,2.60);
  \draw[->, line width=0.8pt, draw=domainmedia!75!black] (10.35,5.85) -- (10.75,5.85);
  \draw[->, line width=0.8pt, draw=domainmedia!75!black] (10.35,2.40) -- (10.75,2.40);
\end{tikzpicture}
}
\refstepcounter{figure}\label{fig:jobbench:reporters_task1}
\vspace{0.35em}
\parbox{\textwidth}{\small\textbf{Figure~\thefigure:} Task schematic for Reporters. The agent receives an editor's assignment and a heterogeneous evidence pack of water-quality CSVs, state surveillance reports, regulatory documents, and interview notes, and must deliver a pitch memo, a workbook of cross-source analyses, and a 15-entry source log.}
\end{center}

\taskinstructionbox[title=Instruction]{appendix-task-examples/reporters-task1-instruction.txt}

\input{appendix-task-examples/reporters-task1-rubrics}

\clearpage
\input{appendix-task-examples/reporters-task1-task-card}

\clearpage
\subsection{Data Entry Keyers: Q3 Sales Order Reconciliation}
\label{app:task-example:data-entry}

\begin{center}
\centering
\resizebox{\textwidth}{!}{\begin{tikzpicture}[font=\sffamily]
  \draw[rounded corners=7pt, line width=1.2pt, draw=domainadmin!90!black, fill=domainadmin!8] (0,0) rectangle (16,8.6);
  \fill[domainadmin!22] (0,7.65) rectangle (16,8.6);
  \node[anchor=west, font=\Large\bfseries, text=domainadmin!90!black] at (0.35,8.13) {Data Entry Keyers -- Task 1};
  \node[anchor=east, font=\small\bfseries, text=domainadmin!80!black] at (15.65,8.13) {Three-way order reconciliation};

  \draw[rounded corners=4pt, draw=jbblue!80!black, fill=white, line width=0.7pt] (0.45,4.45) rectangle (5.25,7.35);
  \fill[jbblue!12] (0.45,6.88) rectangle (5.25,7.35);
  \node[anchor=west, font=\bfseries\small, text=jbblue!85!black] at (0.68,7.12) {Assignment Brief};
  \node[anchor=north west, align=left, text width=4.35cm, font=\scriptsize] at (0.70,6.72) {Act as a senior data entry keyer before a quarterly compliance audit. Reconcile Q3 2025 sales orders across CRM exports, scanned handwritten forms, and ERP records using the specified source hierarchy.};

  \draw[rounded corners=4pt, draw=jbgreen!75!black, fill=white, line width=0.7pt] (0.45,0.55) rectangle (5.25,4.12);
  \fill[jbgreen!13] (0.45,3.65) rectangle (5.25,4.12);
  \node[anchor=west, font=\bfseries\small, text=jbgreen!70!black] at (0.68,3.89) {Evidence Files};
  \node[anchor=north west, align=left, text width=4.35cm, font=\scriptsize] at (0.70,3.45) {
    CRM sales-order workbook\\
    ERP data-entry workbook\\
    10 scanned order forms\\
    Customer master list\\
    Error-log and compliance templates
  };

  \draw[rounded corners=4pt, draw=jborange!90!black, fill=white, line width=0.7pt] (5.65,1.05) rectangle (10.35,7.15);
  \fill[jborange!15] (5.65,6.68) rectangle (10.35,7.15);
  \node[anchor=west, font=\bfseries\small, text=jborange!80!black] at (5.88,6.92) {Reasoning Burden};
  \node[anchor=north west, align=left, text width=4.15cm, font=\scriptsize] at (5.90,6.47) {
    \textbf{Source hierarchy}\\[-0.2em]
    Use scans as authoritative when present, otherwise prioritize CRM over ERP.\\[0.22em]
    \textbf{Reverse calculations}\\[-0.2em]
    Validate units, prices, and totals bidirectionally.\\[0.22em]
    \textbf{Entity validation}\\[-0.2em]
    Check names and countries against the customer master list.\\[0.22em]
    \textbf{Pattern finding}\\[-0.2em]
    Separate systematic errors from random data-entry mistakes.
  };

  \draw[rounded corners=4pt, draw=badgeopenai!75!black, fill=white, line width=0.7pt] (10.75,4.30) rectangle (15.55,7.15);
  \fill[badgeopenai!13] (10.75,6.68) rectangle (15.55,7.15);
  \node[anchor=west, font=\bfseries\small, text=badgeopenai!80!black] at (10.98,6.92) {Required Deliverables};
  \node[anchor=north west, align=left, text width=4.25cm, font=\scriptsize] at (11.00,6.47) {
    \texttt{Discrepancy log}: field-level conflicts, source system, and reconciled values.\\[0.28em]
    \texttt{Reconciled master}: clean Q3 order table ready for audit.\\[0.28em]
    \texttt{Findings report}: systematic error patterns and compliance checklist.
  };

  \draw[rounded corners=4pt, draw=jbbrick!90!black, fill=white, line width=0.7pt] (10.75,0.85) rectangle (15.55,3.90);
  \fill[jbbrick!13] (10.75,3.43) rectangle (15.55,3.90);
  \node[anchor=west, font=\bfseries\small, text=jbbrick!90!black] at (10.98,3.67) {Rubric Pressure Points};
  \node[anchor=north west, align=left, text width=4.25cm, font=\scriptsize] at (11.00,3.22) {Known calculation errors; typo and transposition detection; unregistered customers; country normalization; systematic-vs-random classification; audit-ready reconciled master.};

  \draw[->, line width=0.8pt, draw=domainadmin!75!black] (5.25,5.90) -- (5.65,5.90);
  \draw[->, line width=0.8pt, draw=domainadmin!75!black] (5.25,2.20) -- (5.65,2.60);
  \draw[->, line width=0.8pt, draw=domainadmin!75!black] (10.35,5.85) -- (10.75,5.85);
  \draw[->, line width=0.8pt, draw=domainadmin!75!black] (10.35,2.40) -- (10.75,2.40);
\end{tikzpicture}
}
\refstepcounter{figure}\label{fig:jobbench:data_entry_task1}
\vspace{0.35em}
\parbox{\textwidth}{\small\textbf{Figure~\thefigure:} Task schematic for Data Entry Keyers. The agent receives Q3 sales orders across CRM exports, scanned handwritten forms, and ERP records, and must reconcile them under a specified source hierarchy and deliver a discrepancy log, a clean reconciled master, and a findings report.}
\end{center}

\taskinstructionbox[title=Instruction]{appendix-task-examples/data-entry-task1-instruction.txt}

\input{appendix-task-examples/data-entry-task1-rubrics}

\clearpage
\input{appendix-task-examples/data-entry-task1-task-card}

\clearpage
\subsection{Lawyers: Short-Term Rental Ordinance Settlement Analysis}
\label{app:task-example:lawyers}

\begin{center}
\centering
\resizebox{\textwidth}{!}{\begin{tikzpicture}[font=\sffamily]
  \draw[rounded corners=7pt, line width=1.2pt, draw=domainlegal!90!black, fill=domainlegal!8] (0,0) rectangle (16,8.6);
  \fill[domainlegal!22] (0,7.65) rectangle (16,8.6);
  \node[anchor=west, font=\Large\bfseries, text=domainlegal!90!black] at (0.35,8.13) {Lawyers -- Task 1};
  \node[anchor=east, font=\small\bfseries, text=domainlegal!80!black] at (15.65,8.13) {STR ordinance settlement response};

  \draw[rounded corners=4pt, draw=jbblue!80!black, fill=white, line width=0.7pt] (0.45,4.45) rectangle (5.25,7.35);
  \fill[jbblue!12] (0.45,6.88) rectangle (5.25,7.35);
  \node[anchor=west, font=\bfseries\small, text=jbblue!85!black] at (0.68,7.12) {Assignment Brief};
  \node[anchor=north west, align=left, text width=4.35cm, font=\scriptsize] at (0.70,6.72) {Act as a North Carolina real-estate and land-use attorney. Evaluate a town settlement offer for six short-term-rental properties and prepare constitutional arguments, comparative regulation analysis, and a counter-proposal.};

  \draw[rounded corners=4pt, draw=jbgreen!75!black, fill=white, line width=0.7pt] (0.45,0.55) rectangle (5.25,4.12);
  \fill[jbgreen!13] (0.45,3.65) rectangle (5.25,4.12);
  \node[anchor=west, font=\bfseries\small, text=jbgreen!70!black] at (0.68,3.89) {Evidence Files};
  \node[anchor=north west, align=left, text width=4.35cm, font=\scriptsize] at (0.70,3.45) {
    Town settlement letter\\
    Client property database\\
    Belvidere STR ordinance\\
    Cornell STR regulations PDF\\
    Case-law and Penn Central context
  };

  \draw[rounded corners=4pt, draw=jborange!90!black, fill=white, line width=0.7pt] (5.65,1.05) rectangle (10.35,7.15);
  \fill[jborange!15] (5.65,6.68) rectangle (10.35,7.15);
  \node[anchor=west, font=\bfseries\small, text=jborange!80!black] at (5.88,6.92) {Reasoning Burden};
  \node[anchor=north west, align=left, text width=4.15cm, font=\scriptsize] at (5.90,6.47) {
    \textbf{Legal synthesis}\\[-0.2em]
    Compare local ordinance provisions with relevant STR litigation and doctrine.\\[0.22em]
    \textbf{Financial exposure}\\[-0.2em]
    Quantify fines, revenue loss, permit denials, and settlement pressure.\\[0.22em]
    \textbf{Property-specific analysis}\\[-0.2em]
    Separate facts across six properties rather than arguing generically.\\[0.22em]
    \textbf{Negotiation framing}\\[-0.2em]
    Translate legal and financial analysis into settlement terms.
  };

  \draw[rounded corners=4pt, draw=badgeopenai!75!black, fill=white, line width=0.7pt] (10.75,4.30) rectangle (15.55,7.15);
  \fill[badgeopenai!13] (10.75,6.68) rectangle (15.55,7.15);
  \node[anchor=west, font=\bfseries\small, text=badgeopenai!80!black] at (10.98,6.92) {Required Deliverables};
  \node[anchor=north west, align=left, text width=4.25cm, font=\scriptsize] at (11.00,6.47) {
    \texttt{Legal memo}: constitutional and settlement-risk analysis.\\[0.28em]
    \texttt{Regulatory table}: jurisdiction-by-jurisdiction STR comparison.\\[0.28em]
    \texttt{Counter-proposal}: client-oriented settlement response.
  };

  \draw[rounded corners=4pt, draw=jbbrick!90!black, fill=white, line width=0.7pt] (10.75,0.85) rectangle (15.55,3.90);
  \fill[jbbrick!13] (10.75,3.43) rectangle (15.55,3.90);
  \node[anchor=west, font=\bfseries\small, text=jbbrick!90!black] at (10.98,3.67) {Rubric Pressure Points};
  \node[anchor=north west, align=left, text width=4.25cm, font=\scriptsize] at (11.00,3.22) {Fine exposure; seven-category regulatory table; owner-occupancy and permit caps; Penn Central factors; property-by-property financials; viable counterterms.};

  \draw[->, line width=0.8pt, draw=domainlegal!75!black] (5.25,5.90) -- (5.65,5.90);
  \draw[->, line width=0.8pt, draw=domainlegal!75!black] (5.25,2.20) -- (5.65,2.60);
  \draw[->, line width=0.8pt, draw=domainlegal!75!black] (10.35,5.85) -- (10.75,5.85);
  \draw[->, line width=0.8pt, draw=domainlegal!75!black] (10.35,2.40) -- (10.75,2.40);
\end{tikzpicture}
}
\refstepcounter{figure}\label{fig:jobbench:lawyers_task1}
\vspace{0.35em}
\parbox{\textwidth}{\small\textbf{Figure~\thefigure:} Task schematic for Lawyers. The agent receives a town settlement letter, a six-property client database, and a packet of ordinances and case law, and must deliver a constitutional and settlement-risk memo, a jurisdiction-by-jurisdiction regulatory table, and a counter-proposal.}
\end{center}

\taskinstructionbox[title=Instruction]{appendix-task-examples/lawyers-task1-instruction.txt}

\input{appendix-task-examples/lawyers-task1-rubrics}

\clearpage
\input{appendix-task-examples/lawyers-task1-task-card}

\clearpage
\subsection{Web Administrators: ShopVault Incident Reconstruction}
\label{app:task-example:web-admin}

\begin{center}
\centering
\resizebox{\textwidth}{!}{\begin{tikzpicture}[font=\sffamily]
  \draw[rounded corners=7pt, line width=1.2pt, draw=domainit!90!black, fill=domainit!8] (0,0) rectangle (16,8.6);
  \fill[domainit!22] (0,7.65) rectangle (16,8.6);
  \node[anchor=west, font=\Large\bfseries, text=domainit!90!black] at (0.35,8.13) {Web Administrators -- Task 1};
  \node[anchor=east, font=\small\bfseries, text=domainit!80!black] at (15.65,8.13) {Incident reconstruction and hardening};

  \draw[rounded corners=4pt, draw=jbblue!80!black, fill=white, line width=0.7pt] (0.45,4.45) rectangle (5.25,7.35);
  \fill[jbblue!12] (0.45,6.88) rectangle (5.25,7.35);
  \node[anchor=west, font=\bfseries\small, text=jbblue!85!black] at (0.68,7.12) {Assignment Brief};
  \node[anchor=north west, align=left, text width=4.35cm, font=\scriptsize] at (0.70,6.72) {Act as a web administrator investigating ShopVault after a hardening change request was closed without verification. Reconstruct attack chains from logs, audit configs, and produce hardened nginx and firewall rules.};

  \draw[rounded corners=4pt, draw=jbgreen!75!black, fill=white, line width=0.7pt] (0.45,0.55) rectangle (5.25,4.12);
  \fill[jbgreen!13] (0.45,3.65) rectangle (5.25,4.12);
  \node[anchor=west, font=\bfseries\small, text=jbgreen!70!black] at (0.68,3.89) {Evidence Files};
  \node[anchor=north west, align=left, text width=4.35cm, font=\scriptsize] at (0.70,3.45) {
    Access logs and threat indicators\\
    Change request CR-2025-0347\\
    nginx.conf and iptables.rules\\
    Network topology JSON\\
    SSL audit and incident template
  };

  \draw[rounded corners=4pt, draw=jborange!90!black, fill=white, line width=0.7pt] (5.65,1.05) rectangle (10.35,7.15);
  \fill[jborange!15] (5.65,6.68) rectangle (10.35,7.15);
  \node[anchor=west, font=\bfseries\small, text=jborange!80!black] at (5.88,6.92) {Reasoning Burden};
  \node[anchor=north west, align=left, text width=4.15cm, font=\scriptsize] at (5.90,6.47) {
    \textbf{Kill-chain reconstruction}\\[-0.2em]
    Sequence SQLi, brute force, IDOR, XSS, traversal, and destructive actions.\\[0.22em]
    \textbf{Compliance audit}\\[-0.2em]
    Map missed change-request controls to observed exploit outcomes.\\[0.22em]
    \textbf{Topology-aware repair}\\[-0.2em]
    Respect zone CIDRs when hardening nginx and iptables.\\[0.22em]
    \textbf{Incident reporting}\\[-0.2em]
    Convert technical evidence into NIST-style findings and actions.
  };

  \draw[rounded corners=4pt, draw=badgeopenai!75!black, fill=white, line width=0.7pt] (10.75,4.30) rectangle (15.55,7.15);
  \fill[badgeopenai!13] (10.75,6.68) rectangle (15.55,7.15);
  \node[anchor=west, font=\bfseries\small, text=badgeopenai!80!black] at (10.98,6.92) {Required Deliverables};
  \node[anchor=north west, align=left, text width=4.25cm, font=\scriptsize] at (11.00,6.47) {
    \texttt{findings\_summary.csv}: attack-chain events with phases and success flags.\\[0.28em]
    \texttt{cr\_compliance\_audit.csv}: control gaps tied to outcomes.\\[0.28em]
    \texttt{hardened configs}: nginx, iptables, and incident report.
  };

  \draw[rounded corners=4pt, draw=jbbrick!90!black, fill=white, line width=0.7pt] (10.75,0.85) rectangle (15.55,3.90);
  \fill[jbbrick!13] (10.75,3.43) rectangle (15.55,3.90);
  \node[anchor=west, font=\bfseries\small, text=jbbrick!90!black] at (10.98,3.67) {Rubric Pressure Points};
  \node[anchor=north west, align=left, text width=4.25cm, font=\scriptsize] at (11.00,3.22) {Full kill-chain timeline; CR gap-to-outcome mapping; dotfile/debug/admin-control fixes; topology-safe firewall rules; TLS findings; NIST incident report completeness.};

  \draw[->, line width=0.8pt, draw=domainit!75!black] (5.25,5.90) -- (5.65,5.90);
  \draw[->, line width=0.8pt, draw=domainit!75!black] (5.25,2.20) -- (5.65,2.60);
  \draw[->, line width=0.8pt, draw=domainit!75!black] (10.35,5.85) -- (10.75,5.85);
  \draw[->, line width=0.8pt, draw=domainit!75!black] (10.35,2.40) -- (10.75,2.40);
\end{tikzpicture}
}
\refstepcounter{figure}\label{fig:jobbench:web_admin_task1}
\vspace{0.35em}
\parbox{\textwidth}{\small\textbf{Figure~\thefigure:} Task schematic for Web Administrators. The agent receives access logs, threat indicators, a closed change request, live nginx and iptables configuration, and a network topology, and must reconstruct the kill chain, audit the missed controls, and deliver hardened configurations and a NIST-style incident report.}
\end{center}

\taskinstructionbox[title=Instruction]{appendix-task-examples/web-admin-task1-instruction.txt}

\input{appendix-task-examples/web-admin-task1-rubrics}

\clearpage
\input{appendix-task-examples/web-admin-task1-task-card}

%% file: appendix-task-examples/reporters-task1-rubrics.tex
\begin{taskrubricreadablebox}[title={Rubrics (8 checks, 60 total points)}]
\begin{enumerate}[leftmargin=1.35em,itemsep=0.55em,topsep=0.25em]
\item \textbf{8 pts.} Does the response correctly map which of the five CSV water system cities appear on the CT surveillance report's 2024 high-risk communities list, and which do not?
\begin{itemize}[leftmargin=1.1em,itemsep=0.12em,topsep=0.15em]
\item The response must correctly state that all five water system cities from the CSV (Hartford, New Haven, Bridgeport, Waterbury, Meriden) appear on the CT 2024 Surveillance Report's top 10 high-risk communities list, with their rankings: New Haven (\#1), Bridgeport (\#2), Waterbury (\#3), Hartford (\#4), Meriden (\#5).
\item The response must explain the significance of the high-risk designation -- that it is based on rates of newly poisoned children, age of housing, and income levels, not water lead levels specifically.
\item The response must acknowledge that the remaining 5 high-risk communities (New Britain, Norwich, West Haven, Stamford, Manchester) are not represented in the water testing CSV data, representing a potential gap in the story's coverage.
\end{itemize}
\item \textbf{8 pts.} Does the data analysis workbook's 'Water Lead Trends' sheet correctly compute the percentage change in 90th-percentile lead levels from 2020 to 2024 for each of the five water systems?
\begin{itemize}[leftmargin=1.1em,itemsep=0.12em,topsep=0.15em]
\item The workbook must compute percentage changes in 90th-percentile lead levels from 2020 to 2024 for all five water systems, with approximately correct values: Hartford \textasciitilde{}+26.8\% (8.2 to 10.4 ppb), New Haven \textasciitilde{}+14.1\% (6.4 to 7.3 ppb), Bridgeport \textasciitilde{}+15.7\% (5.1 to 5.9 ppb), Waterbury \textasciitilde{}+31.6\% (9.8 to 12.9 ppb), Meriden \textasciitilde{}+41.9\% (4.3 to 6.1 ppb).
\item Alternative valid calculation methodologies (e.g., using 2020 averages or 2024 averages instead of single periods) are acceptable, but the methodology must be clearly stated and the math must be correct.
\item The workbook should flag which systems exceed 15 ppb and 10 ppb thresholds across which monitoring periods.
\end{itemize}
\item \textbf{10 pts.} Does the 'CT vs National' sheet correctly extract Connecticut's data from the CDC dataset and compare it to national totals, noting that CT data is only available for 2017-2019?
\begin{itemize}[leftmargin=1.1em,itemsep=0.12em,topsep=0.15em]
\item The sheet must note that Connecticut data in the CDC dataset is only available for 2017, 2018, and 2019, with years 2020-2022 marked as not submitted.
\item The sheet must present correct figures: 2017 CT had 1,666 tested with 93 (5.6\%) at >=5 g/dL vs. national 2,580,144 tested with 38,427 (1.5\%); 2018 CT had 19,946 tested with 112 (0.6\%) vs. national 2,923,721 with 36,990 (1.3\%); 2019 CT had 35,881 tested with 101 (0.3\%) vs. national 2,691,749 with 29,137 (1.1\%).
\item The sheet must note the 2017 CT anomaly (only 1,666 tested, yielding an artificially high 5.6\% percentage) and explain that the BLL >=3.5 threshold data is only available nationally for 2022 (31,767 or 1.7\%) and is unavailable for CT in the CDC dataset.
\end{itemize}
\item \textbf{6 pts.} Does the 'City Cross-Reference' sheet correctly map each water system city to the town-level blood lead data from the CT surveillance report, including total confirmed tests, and counts at each BLL threshold?
\begin{itemize}[leftmargin=1.1em,itemsep=0.12em,topsep=0.15em]
\item The sheet must include correct town-level data for all five cities: Hartford (3,539 total, 150 at >=3.5, 92 at >=5, 29 at >=10, 12 at >=15, 7 at >=20 g/dL), New Haven (3,185 total, 224 at >=3.5, 103 at >=5, 21 at >=10, 5 at >=15, 2 at >=20), Bridgeport (4,188 total, 215 at >=3.5, 113 at >=5, 29 at >=10, 14 at >=15, 5 at >=20), Waterbury (3,662 total, 196 at >=3.5, 92 at >=5, 23 at >=10, 15 at >=15, 9 at >=20), and Meriden (1,923 total, 71 at >=3.5, 29 at >=5, 6 at >=10, 2 at >=15, 2 at >=20).
\item The sheet must include the high-risk rankings for each city: New Haven (\#1), Bridgeport (\#2), Waterbury (\#3), Hartford (\#4), Meriden (\#5).
\item Significant errors in the town-level numbers (beyond minor rounding) fail this rubric.
\end{itemize}
\item \textbf{8 pts.} Does the source verification log contain at least 15 entries and use the correct CSV format with columns Data\_Point, Source\_File, Page\_or\_Location, Verified (Yes/No/Partial), and Notes?
\begin{itemize}[leftmargin=1.1em,itemsep=0.12em,topsep=0.15em]
\item The source log must be a CSV file with exactly five columns: Data\_Point, Source\_File, Page\_or\_Location, Verified (Yes/No/Partial), and Notes.
\item The log must contain at least 15 distinct entries covering key statistics, quotes, and claims used in the pitch memo and data workbook.
\item Entries must span multiple source files -- at minimum including entries from: the CSV water data, the CT surveillance report, the CDC national data, the EPA LCRI factsheet, the NRDC Newark factsheet, and Dr. Martinez's interview.
\item Each entry must have a specific page/location reference (e.g., 'CT report p.3', 'CSV row 38', 'EPA factsheet p.1') rather than vague references like 'report'.
\end{itemize}
\item \textbf{6 pts.} Does the response reference the correct legislative basis for Connecticut's lowered blood lead reference value -- Public Act 22-49, effective January 1, 2023, lowering the threshold from 5 g/dL to 3.5 g/dL?
\begin{itemize}[leftmargin=1.1em,itemsep=0.12em,topsep=0.15em]
\item The response must correctly identify Connecticut Public Act 22-49 as the legislative basis for the lowered blood lead reference value.
\item The response must state the correct effective date (January 1, 2023) and the correct threshold change (from 5 g/dL to 3.5 g/dL).
\item Attributing the change to the wrong act number, wrong date, or wrong threshold values fails this rubric.
\end{itemize}
\item \textbf{8 pts.} Does the pitch memo discuss the lead service line counts and replacement burden across the five water systems, connecting this to the LCRI's 10-year replacement mandate?
\begin{itemize}[leftmargin=1.1em,itemsep=0.12em,topsep=0.15em]
\item The pitch memo must include lead service line counts and population served for each system from the 2024 CSV data: Hartford (2,500 lines, 400,000 people, 0.63\% ratio), New Haven (1,250 lines, 430,000 people, 0.29\%), Bridgeport (1,550 lines, 350,000 people, 0.44\%), Waterbury (1,180 lines, 220,000 people, 0.54\%), Meriden (650 lines, 120,000 people, 0.54\%).
\item The pitch memo must identify Hartford as having both the most absolute lead service lines (2,500) and the highest ratio of lines to population served (0.63\%).
\item The pitch memo must connect these service line counts to the LCRI's requirement that systems replace lead service lines within 10 years, identifying which systems face the largest replacement burdens.
\end{itemize}
\item \textbf{6 pts.} Does the data analysis workbook flag 90th-percentile readings that exceed both the 15 ppb and 10 ppb thresholds for each monitoring period, correctly identifying all exceedance periods?
\begin{itemize}[leftmargin=1.1em,itemsep=0.12em,topsep=0.15em]
\item The workbook must correctly identify the 15 ppb exceedances: Waterbury Jan-Jun 2023 (15.3 ppb) and Jul-Dec 2023 (16.1 ppb).
\item The workbook must correctly identify periods exceeding 10 ppb but below 15 ppb, including: Hartford Jan-Jun 2022 (11.3), Jul-Dec 2022 (12.7), Jan-Jun 2023 (13.2), Jul-Dec 2023 (14.1), Jan-Jun 2024 (11.8), Jul-Dec 2024 (10.4); and Waterbury Jul-Dec 2020 (10.4), Jan-Jun 2021 (11.2), Jul-Dec 2021 (12.1), Jan-Jun 2022 (13.5), Jul-Dec 2022 (14.8), Jan-Jun 2024 (14.2), Jul-Dec 2024 (12.9).
\item The workbook must clearly distinguish which periods exceed the 15 ppb current action level versus which exceed only the 10 ppb LCRI threshold.
\item The CSV's 'Action\_Level\_Exceedance' column marks Hartford's 2024 periods and Waterbury's 2023-2024 periods as 'Yes' (under the current 15 ppb standard), and the workbook should reflect this.
\end{itemize}
\end{enumerate}
\end{taskrubricreadablebox}

%% file: appendix-task-examples/reporters-task1-task-card.tex
\begin{taskcardreadablebox}[title={Task Card}]
\textbf{Summary.} This task asks the agent to act as an investigative reporter at a Connecticut outlet, building a pre-publication editorial package on lead in drinking water and childhood health: a thesis-driven pitch memo, a three-sheet data analysis workbook, and a 15+ entry source verification log -- all internally cross-checked against the FOIA water data, the CT surveillance report, and an interview with Dr. Martinez.\par
\vspace{0.35em}
\begin{tabularx}{\textwidth}{@{}>{\bfseries}p{0.23\textwidth}Y@{}}
ONET task & Check reference materials, such as books, news files, or public records, to obtain relevant facts. \\
Expert desire & 4.00 (scale: 0--5, from WorkBank) \\
In practice & harvesting verifiable data from heterogeneous reference materials (FOIA CSVs, state PDF reports, NRDC factsheets, EPA regulatory factsheets, federal CDC datasets, interview transcripts), pairing every claim with a sourceable data point, and assembling a fact-tracked analytical package that an editor can defend on the front page. \\
\end{tabularx}
\vspace{0.35em}
\textbf{Why desired.} Investigative beat reporting is gated by source-verification time -- extracting numbers from PDFs, joining FOIA CSVs to surveillance tables, and cross-checking interview claims against public records. An agent that handles the verification layer lets the reporter focus on the editorial argument and the framing.\par
\vspace{0.35em}
\textbf{Reasoning challenges.}
\begin{itemize}[leftmargin=1.25em,itemsep=0.22em,topsep=0.2em]
\item \textbf{Reconciling contradictory water-vs-paint evidence.} The CT 2024 surveillance report shows 0\% of investigated homes identified water as a lead hazard, yet the FOIA CSV shows multiple Hartford-area systems exceeding the 15 ppb federal action level. The agent must reason about home-investigation methodology limits versus system-level monitoring data and produce a coherent editorial argument, not paper over the contradiction.
\item \textbf{Fact-checking Dr. Martinez's "30\% increase" quote.} Dr. Martinez claims a 30\% post-threshold-change increase in referrals. The CT surveillance report shows \textasciitilde{}66\% statewide prevalence increase (2022->2023) and \textasciitilde{}84\% incidence increase (2022->2024). The agent must classify the quote as directionally correct but quantitatively conservative -- likely clinic-specific -- rather than accept or reject it outright.
\item \textbf{Waterbury-vs-Newark data parallel.} Waterbury peaked at 16.1 ppb (just above the 15 ppb threshold); Newark hit 47.9 ppb (over 3 x the threshold). The agent must construct a structural-trajectory parallel (sustained exceedances, delayed enforcement, vulnerable populations) rather than a point-in-time comparison that obscures the gap.
\item \textbf{City-to-surveillance mapping with a credibility gap.} All five FOIA water-system service cities appear on the CT surveillance Top-10 high-risk communities list, but five high-risk communities (New Britain, Norwich, West Haven, Stamford, Manchester) are NOT in the water data -- a sourcing gap the agent must flag for story credibility.
\item \textbf{LCRI: finalized vs. enforceable.} The Lead and Copper Rule Improvements were finalized October 2024, but the new 10 ppb action level does not yet apply to current monitoring cycles. The agent must read the EPA factsheet carefully and distinguish "finalized" from "in enforcement" -- the distinction the editor explicitly flagged.
\item \textbf{CDC data submission gaps for Connecticut.} CDC childhood blood lead surveillance for CT is only available 2017--2019; 2020--2022 are dagger-marked non-submissions. The 2017 row covers only 1,666 children tested, producing an artificially elevated 5.6\% rate that the agent must flag for sampling bias.
\item \textbf{Counterintuitive racial-disparity ordering.} The CT report shows Non-Hispanic Asian children (4.0\%) and Hispanic children (4.0\%) above Non-Hispanic Black children (2.9\%) in incidence. The agent must report these accurately rather than default to common-assumption ordering, and connect them to Dr. Martinez's quote about children of color.
\end{itemize}
\vspace{0.2em}
\textbf{Files requiring search.} External federal data and regulatory factsheets the agent must locate.
\begin{itemize}[leftmargin=1.25em,itemsep=0.22em,topsep=0.2em]
\item \texttt{CDC\_2017\_2022\_Blood\_Lead\_National\_Data.xlsx}: State-level CDC childhood blood lead surveillance data -- provides Connecticut 2017--2019 rows and national totals for the "CT vs National" sheet. \emph{Source:} cdc.gov (\url{https://www.cdc.gov/lead-prevention/media/files/2025/08/2017-2022-cbls-national-data-508-1.xlsx}).
\item \texttt{EPA\_LCRI\_General\_Public\_Factsheet.pdf}: EPA factsheet on the Lead and Copper Rule Improvements -- supplies the finalized 10 ppb action level, 10-year service-line replacement mandate, and enforcement timing needed to discuss current vs. forthcoming regulation accurately. \emph{Source:} epa.gov (\url{https://www.epa.gov/system/files/documents/2024-10/final_lcri_fact-sheet_general_public.pdf}).
\end{itemize}
\end{taskcardreadablebox}

%% file: appendix-task-examples/data-entry-task1-rubrics.tex
\begin{taskrubricreadablebox}[title={Rubrics (12 checks, 92 total points)}]
\begin{enumerate}[leftmargin=1.35em,itemsep=0.55em,topsep=0.25em]
\item \textbf{10 pts.} Does the Error Log correctly document the calculation error for ORD-202507-0002 where ERP shows Total\_Revenue of \$6794.25 instead of the correct \$6749.25?
\begin{itemize}[leftmargin=1.1em,itemsep=0.12em,topsep=0.15em]
\item The Error Log must document that for ORD-202507-0002, ERP shows Total\_Revenue as \$6794.25 while the correct calculation is 75  x  \$89.99 = \$6749.25 (matching CRM and scanned form).
\item The error type must be classified as 'Calculation' with the source system identified as 'ERP'.
\item The reconciled value must be recorded as \$6749.25.
\item Root cause notes should show the expected formula (75  x  \$89.99) and the \$45.00 deviation.
\end{itemize}
\item \textbf{6 pts.} Does the Error Log document the customer name typo for ORD-202507-0004 where ERP has 'Pacific Rim Export' instead of 'Pacific Rim Exports'?
\begin{itemize}[leftmargin=1.1em,itemsep=0.12em,topsep=0.15em]
\item The Error Log must document that for ORD-202507-0004, ERP has Customer\_Name as 'Pacific Rim Export' (missing 's') while CRM and scanned form have 'Pacific Rim Exports'.
\item The error type must be classified as 'Typo' with the source system identified as 'ERP'.
\item The reconciled value must be 'Pacific Rim Exports', confirmed by Customer\_Master\_List.csv where CUST-004 is registered as 'Pacific Rim Exports'.
\end{itemize}
\item \textbf{6 pts.} Does the Error Log identify the digit transposition error for ORD-202507-0005 where ERP shows 1020 units sold instead of the correct 1200 units?
\begin{itemize}[leftmargin=1.1em,itemsep=0.12em,topsep=0.15em]
\item The Error Log must document that for ORD-202507-0005, ERP shows Units\_Sold as 1020 while CRM and scanned form show 1200.
\item The error type must be classified as 'Transposition' (digits 0 and 2 swapped) with the source system identified as 'ERP'.
\item The reconciled value must be 1200.
\end{itemize}
\item \textbf{10 pts.} Does the Error Log correctly flag the country naming inconsistency for ORD-202508-0006 where CRM shows 'Italia' instead of 'Italy'?
\begin{itemize}[leftmargin=1.1em,itemsep=0.12em,topsep=0.15em]
\item The Error Log must document that for ORD-202508-0006, CRM shows Country as 'Italia' while ERP and scanned form show 'Italy'.
\item The Error Log must reference Customer\_Master\_List.csv to confirm CUST-006 (Mediterranean Trade Co) is registered with Country 'Italy'.
\item The error type must be classified as 'Format' (non-standard naming) with the source system identified as 'CRM'.
\item The reconciled value must be 'Italy'.
\end{itemize}
\item \textbf{10 pts.} Does the Error Log identify the Unit Price discrepancy for ORD-202508-0008 where ERP shows \$5.05 instead of \$5.50?
\begin{itemize}[leftmargin=1.1em,itemsep=0.12em,topsep=0.15em]
\item The Error Log must document that for ORD-202508-0008, ERP shows Unit\_Price as \$5.05 while CRM and scanned form show \$5.50.
\item The log should note that ERP Total\_Revenue of \$4400.00 matches 800  x  \$5.50, indicating the ERP Unit\_Price field is internally inconsistent with its own revenue.
\item The error type must be classified as 'Typo' with the source system identified as 'ERP'.
\item The reconciled value must be \$5.50.
\end{itemize}
\item \textbf{6 pts.} Does the Error Log correctly identify the date error for ORD-202509-0009 where CRM shows 09/30/2025 but ERP and scanned form show 09/03/2025?
\begin{itemize}[leftmargin=1.1em,itemsep=0.12em,topsep=0.15em]
\item The Error Log must document that for ORD-202509-0009, CRM shows Order\_Date as '09/30/2025' while ERP shows '09/03/2025' and scanned form confirms '09/03/2025'.
\item Per source priority (scanned highest), the reconciled date must be 09/03/2025.
\item The error type must be classified as 'Transposition' (day digits 03 vs 30 swapped) with the source system identified as 'CRM'.
\end{itemize}
\item \textbf{6 pts.} Does the Error Log document the country name typo for ORD-202509-0010 where ERP has 'Canda' instead of 'Canada'?
\begin{itemize}[leftmargin=1.1em,itemsep=0.12em,topsep=0.15em]
\item The Error Log must document that for ORD-202509-0010, ERP has Country as 'Canda' while CRM and scanned form show 'Canada'.
\item The Error Log must reference Customer\_Master\_List.csv to confirm CUST-010 (Canadian Wholesale Inc) is registered with Country 'Canada'.
\item The error type must be classified as 'Typo' with the source system identified as 'ERP' and the reconciled value as 'Canada'.
\end{itemize}
\item \textbf{6 pts.} Does the Error Log identify the Item Type typo for ORD-202508-0018 where ERP has 'Houshold' instead of 'Household'?
\begin{itemize}[leftmargin=1.1em,itemsep=0.12em,topsep=0.15em]
\item The Error Log must document that for ORD-202508-0018, ERP has Item\_Type as 'Houshold' (missing 'e') while CRM shows 'Household'.
\item The error type must be classified as 'Typo' with the source system identified as 'ERP'.
\item The reconciled value must be 'Household'.
\end{itemize}
\item \textbf{6 pts.} Does the Error Log identify the Unit Price transposition for ORD-202509-0023 where ERP shows \$108.00 instead of \$180.00?
\begin{itemize}[leftmargin=1.1em,itemsep=0.12em,topsep=0.15em]
\item The Error Log must document that for ORD-202509-0023, ERP shows Unit\_Price as \$108.00 while CRM shows \$180.00.
\item The Error Log must use Total\_Revenue of \$8100.00 to validate the CRM price (45  x  \$180.00 = \$8100.00), not the ERP price (45  x  \$108.00 = \$4860.00).
\item The error type must be classified as 'Transposition' with the source system identified as 'ERP'.
\item The reconciled value must be \$180.00.
\end{itemize}
\item \textbf{6 pts.} Does the Error Log document the Sales Channel discrepancy for ORD-202509-0026 where ERP shows 'Online' but CRM shows 'Offline'?
\begin{itemize}[leftmargin=1.1em,itemsep=0.12em,topsep=0.15em]
\item The Error Log must document that for ORD-202509-0026, ERP has Sales\_Channel as 'Online' while CRM shows 'Offline'.
\item Since no scanned form exists for this order, CRM takes priority; the source system causing the error must be identified as 'ERP' with the reconciled value as 'Offline'.
\item The error pattern should be classified as 'Systematic' as multiple Sales Channel errors from ERP exist (e.g., ORD-202508-0007).
\end{itemize}
\item \textbf{10 pts.} Does the Audit Findings Report include a data quality score calculated as percentage of records with no discrepancies?
\begin{itemize}[leftmargin=1.1em,itemsep=0.12em,topsep=0.15em]
\item The Audit\_Findings\_Report.txt must include a data quality score calculated as (records with no discrepancies / total records)  x  100\%.
\item The calculation must be based on 30 total records, with discrepancies identified in all 17 of the following orders: 0001-0010, 0012, 0015, 0018, 0020, 0023, 0026, 0029.
\item The data quality score should accurately reflect the count of clean vs. discrepant records.
\item The report should clearly state the methodology used for the data quality score calculation.
\end{itemize}
\item \textbf{10 pts.} Does the Compliance Checklist correctly assess CRT-005 (Total Revenue Calculation Accuracy) as 'Fail'?
\begin{itemize}[leftmargin=1.1em,itemsep=0.12em,topsep=0.15em]
\item The Compliance\_Checklist.xlsx must mark CRT-005 'Total Revenue calculations are accurate (Units x Unit Price)' as 'Fail'.
\item The Compliance\_Checklist.xlsx must justify the failure by referencing ORD-202507-0002 (ERP: \$6794.25 vs correct \$6749.25) having a calculation error exceeding the +/-\$0.01 tolerance.
\item The Compliance\_Checklist.xlsx must also reference ORD-202508-0015 (ERP: \$4480.00 vs correct \$4880.00) as an additional calculation error exceeding the +/-\$0.01 tolerance.
\end{itemize}
\end{enumerate}
\end{taskrubricreadablebox}

%% file: appendix-task-examples/data-entry-task1-task-card.tex
\begin{taskcardreadablebox}[title={Task Card}]
\textbf{Summary.} This task asks the agent to act as Senior Data Entry Keyer at Global Trade Solutions, three-way reconciling Q3 2025 sales orders across the CRM export, ten scanned handwritten order forms, and the ERP entry records ahead of a quarterly compliance audit -- producing a discrepancy log, a reconciled master, an executive findings report, and a completed compliance checklist.\par
\vspace{0.35em}
\begin{tabularx}{\textwidth}{@{}>{\bfseries}p{0.23\textwidth}Y@{}}
ONET task & Compare data with source documents, or re-enter data in verification format to detect errors. \\
Expert desire & 3.50 (scale: 0--5, from WorkBank) \\
In practice & OCR-reading the ten scanned PNG order forms (orders 0001--0010), comparing them against \texttt{Source\_Sales\_Orders\_Q3\_2025.xlsx} and \texttt{Data\_Entry\_Records\_Q3\_2025.xlsx}, validating customer entities against the 35 registered customers in \texttt{Customer\_Master\_List.csv}, applying source priority (scans > CRM > ERP) to each conflict, and tagging each error as systematic vs random in the error log template. \\
\end{tabularx}
\vspace{0.35em}
\textbf{Why desired.} A keyer's pre-audit day is dominated by line-by-line comparison across three formats and by mechanical checks like Total = Units x Price and country-name normalization -- not the audit judgment. Automating the three-way diff, the calculation reverse-validation, and the master-list cross-check lets the keyer focus on the systematic-pattern findings and the process-improvement recommendations the audit committee actually reads.\par
\vspace{0.35em}
\textbf{Reasoning challenges.}
\begin{itemize}[leftmargin=1.25em,itemsep=0.22em,topsep=0.2em]
\item \textbf{Three-way reconciliation with missing sources.} Only orders 0001--0010 have scanned originals. For orders 0011--0030 the agent must fall back to CRM-vs-ERP with CRM as priority, applying the source hierarchy correctly when the authoritative source is absent.
\item \textbf{Reverse-validation of calculation errors.} Some records have correct Total\_Revenue but wrong Unit\_Price or Units\_Sold. The agent must apply Total = Units x Price bidirectionally to identify which component field is in error rather than flagging the total.
\item \textbf{Unregistered customer detection.} Two orders (ORD-202507-0021 and ORD-202508-0022) name "Unknown Supplier Co" and "Mystery Trading LLC" -- neither in the 35-customer master. The agent must cross-reference all 30 orders against the master to surface them.
\item \textbf{Country naming standardization.} "Italia" vs "Italy" is only catchable by anchoring against the registered country list in \texttt{Customer\_Master\_List.csv}, not by a CRM-vs-ERP diff alone.
\item \textbf{Systematic vs random error classification.} The agent must aggregate individual findings into pattern categories -- transposition errors (IDs, digits, dates), calculation errors, Sales Channel encoding errors, typos -- and report frequencies, not just a flat list of discrepancies.
\end{itemize}
\vspace{0.2em}
\textbf{Files requiring search.} None -- all required references are provided in \texttt{task\_folder/}.
\end{taskcardreadablebox}

%% file: appendix-task-examples/lawyers-task1-rubrics.tex
\begin{taskrubricreadablebox}[title={Rubrics (8 checks, 48 total points)}]
\begin{enumerate}[leftmargin=1.35em,itemsep=0.55em,topsep=0.25em]
\item \textbf{6 pts.} Does the analysis correctly identify and quantify the fine exposure for each of the three properties accruing \$500/day fines, using the database data showing fines started accruing on November 29, 2025, and correctly project ongoing exposure through the January 31, 2026 fine-waiver deadline?
\begin{itemize}[leftmargin=1.1em,itemsep=0.12em,topsep=0.15em]
\item Identifies the three specific properties accruing fines at \$500/day: Property 1 (28 Oceanfront Lane), Property 3 (15 Dune Ridge Court), and Property 5 (203 Lighthouse Road)
\item Calculates that as of December 10, 2025, each property had accrued 12 days  x  \$500 = \$6,000 in fines
\item Projects forward that if settlement is not reached by January 31, 2026, fines will have accrued for 64 days per property totaling \$32,000 per property or \$96,000 aggregate
\item Notes that the fine-waiver is conditioned on executing a settlement agreement by January 31, 2026, creating significant time pressure
\end{itemize}
\item \textbf{6 pts.} Does the comparative regulatory analysis table include at minimum the seven required provision categories (owner-occupancy requirements, permit caps, night limits, spacing rules, amortization/grandfathering, fine structures, and zone restrictions) across all three jurisdictions (Millbrook Ordinance 2025-14, Belvidere Township, and New Orleans per Hignell-Stark)?
\begin{itemize}[leftmargin=1.1em,itemsep=0.12em,topsep=0.15em]
\item The CSV table includes all seven required provision categories: owner-occupancy requirements, permit caps, night limits, spacing rules, amortization/grandfathering, fine structures, and zone restrictions
\item Accurately compares provisions across all three jurisdictions (Millbrook Ordinance 2025-14, Belvidere Township, and New Orleans per Hignell-Stark)
\item Owner-occupancy data is accurate: Millbrook requires primary residence/owner-occupancy; Belvidere requires only a local-agent within 45 miles (no owner-occupancy); New Orleans required homestead exemption (found facially discriminatory under dormant Commerce Clause)
\item Night limits, permit caps, and spacing data is accurate: Millbrook caps at 90 nights/year (120 transitional), 25 town-wide permits, 1,000-ft spacing; Belvidere has no annual night cap (27-night max per stay), no permit cap, no spacing; New Orleans had none referenced
\item Amortization, fines, and zone data is accurate: Millbrook 180 days with no grandfathering and \$500/day fines, R-1 prohibition; Belvidere 30-day grace period with escalating fines (\$100/\$500/\$1,500 plus \$250/day), all zones allowed; New Orleans distinguished residential from non-residential
\end{itemize}
\item \textbf{6 pts.} Does the counter-proposal include a request to extend the transitional period beyond the proposed 3 years (ending December 1, 2028), with economic justification grounded in the financial data?
\begin{itemize}[leftmargin=1.1em,itemsep=0.12em,topsep=0.15em]
\item Counter-proposal argues for extending the transitional period beyond the proposed 3 years (ending December 1, 2028)
\item Justifies extension with reference to clients' substantial mortgage balances (ranging from \$198,000 to \$385,000) and property-specific investment-backed expectations based on years of established STR operations (earliest permit from 2017)
\item Suggests a specific longer period (e.g., 5-7 years) with justification
\item References the financial data showing the gap between STR income and LTR income (e.g., Property 3 earned \textasciitilde{}\$67,860/year net from STR vs. \$33,600/year gross from LTR)
\end{itemize}
\item \textbf{6 pts.} Does the counter-proposal include a request for a mutual tolling agreement on the statute of limitations during settlement negotiations?
\begin{itemize}[leftmargin=1.1em,itemsep=0.12em,topsep=0.15em]
\item Counter-proposal includes a request for a mutual tolling agreement that pauses the running of applicable statutes of limitations on the clients' potential constitutional claims (Takings, Due Process, dormant Commerce Clause under 42 U.S.C.  1983)
\item Provides rationale that the January 5, 2026 response deadline and January 31, 2026 fine-waiver deadline create time pressure that could force premature settlement
\item Notes that the Town's proposal requires dismissal "with prejudice," meaning if negotiations fail after limitations period expires, clients lose their litigation option
\item Explains that a tolling agreement protects both parties by allowing good-faith negotiation without the pressure of expiring claims
\end{itemize}
\item \textbf{6 pts.} Does the analysis apply the Penn Central regulatory takings factors (economic impact, investment-backed expectations, and character of government action) to the clients' specific circumstances to assess litigation strength?
\begin{itemize}[leftmargin=1.1em,itemsep=0.12em,topsep=0.15em]
\item Applies the economic impact factor: the ordinance eliminates clients' primary income stream, with historical STR net income ranging from \$47,130 to \$67,860 annually versus LTR gross income of \$21,000 to \$33,600 (roughly 50\%+ income reduction), though properties retain residual value for other uses
\item Applies the investment-backed expectations factor: clients purchased properties between 2016-2021 specifically for or relying on STR operations, held valid permits under prior regime (earliest from 2017), and the regulatory climate before the 2023 study commission did not signal imminent prohibition
\item Applies the character of government action factor: the ordinance is a broad public-welfare regulation but the 180-day amortization with no grandfathering and \$500/day fines create a punitive character; retroactive elimination of previously permitted use weighs toward finding a taking
\item Explicitly identifies and names all three Penn Central factors being applied
\end{itemize}
\item \textbf{6 pts.} Does the analysis identify the dormant Commerce Clause vulnerability in Millbrook's owner-occupancy requirement by drawing on the Hignell-Stark holding that New Orleans' similar residency/homestead requirement was facially discriminatory against interstate commerce?
\begin{itemize}[leftmargin=1.1em,itemsep=0.12em,topsep=0.15em]
\item Identifies that Ordinance 2025-14's owner-occupancy requirement (primary residence) is analogous to New Orleans' homestead exemption requirement held facially discriminatory against interstate commerce in Hignell-Stark
\item Notes the Fifth Circuit found such requirements forbid out-of-state property owners from participating in the STR market in residential zones and applied strict scrutiny (not Pike balancing), under which the law was "virtually per se invalid"
\item Notes that reasonable nondiscriminatory alternatives existed (enforcement, penalties, night caps, permit caps) as the court found
\item Identifies this as a significant legal weakness in the Town's position that strengthens the clients' litigation posture, noting the Fourth Circuit has not yet addressed this but the Fifth Circuit reasoning is persuasive authority
\end{itemize}
\item \textbf{6 pts.} Does the settlement evaluation correctly identify that the Town's settlement requires dismissal of all claims 'with prejudice' and analyze the strategic implications of this term?
\begin{itemize}[leftmargin=1.1em,itemsep=0.12em,topsep=0.15em]
\item Flags that under the "Mutual Terms" section, clients would be required to "dismiss any pending or contemplated legal claims against the Town with prejudice"
\item Explains that dismissal with prejudice permanently extinguishes the clients' constitutional challenges (Takings, dormant Commerce Clause, Due Process) and cannot be refiled
\item Analyzes strategic significance: the dormant Commerce Clause claim is strong based on Hignell-Stark, and a successful constitutional challenge could invalidate the entire ordinance (not just provide transitional relief)
\item The counter-proposal addresses this by proposing dismissal without prejudice, or alternatively a carve-out preserving the right to challenge future amendments
\end{itemize}
\item \textbf{6 pts.} Does the analysis correctly identify that Millbrook's 180-day amortization period with no grandfathering clause is legally vulnerable when compared to amortization periods upheld in the cases cited by the Town Attorney?
\begin{itemize}[leftmargin=1.1em,itemsep=0.12em,topsep=0.15em]
\item Identifies that the Town Attorney cites AVR, Inc. v. City of St. Louis Park and City of Los Angeles v. Gage as support, but these cases involved different types of uses (adult entertainment, industrial) with different investment profiles than residential STR operations
\item Notes that both cited cases provided longer amortization than Millbrook's 180 days
\item Notes Millbrook provides no grandfathering for established operators with years of history, and under Penn Central the adequacy of amortization must be assessed in light of specific investment-backed expectations -- clients with substantial mortgages cannot reasonably recoup investments in 180 days
\item Contrasts with Belvidere Township's less restrictive approach as further evidence of Millbrook's unreasonableness
\end{itemize}
\end{enumerate}
\end{taskrubricreadablebox}

%% file: appendix-task-examples/lawyers-task1-task-card.tex
\begin{taskcardreadablebox}[title={Task Card}]
\textbf{Summary.} This task asks the agent to act as a North Carolina real estate attorney evaluating a Town's settlement offer on six short-term rental properties, building constitutional counter-arguments and producing a memo, comparative regulatory table, and counter-proposal grounded in case law and per-property financials.\par
\vspace{0.35em}
\begin{tabularx}{\textwidth}{@{}>{\bfseries}p{0.23\textwidth}Y@{}}
ONET task & Study Constitution, statutes, decisions, regulations, and ordinances of quasi-judicial bodies to determine ramifications for cases. \\
Expert desire & 3.17 (scale: 0--5, from WorkBank) \\
In practice & pulling the Hignell-Stark Fifth Circuit opinion and the Penn State Law Review Penn Central article, comparing the Millbrook ordinance against Belvidere's STR rules and the New Orleans scheme, and reconciling all of that with the financial and permitting records sitting in the eight-table client database -- then deriving the constitutional theory that drives the settlement response. \\
\end{tabularx}
\vspace{0.35em}
\textbf{Why desired.} Constitutional litigation work is research-saturated -- attorneys spend hours reading opinions, scholarly articles, and comparative ordinances before writing a single line of advocacy. An agent that handles the doctrinal sourcing, ordinance comparison, and per-property financial roll-up frees the attorney to focus on strategy and the judgment call about whether to settle, counter, or litigate.\par
\vspace{0.35em}
\textbf{Reasoning challenges.}
\begin{itemize}[leftmargin=1.25em,itemsep=0.22em,topsep=0.2em]
\item \textbf{Turning the Town's own cited case against the Town.} The settlement letter cites Hignell-Stark v. New Orleans (5th Cir. 2022) defensively, but the same opinion's dormant Commerce Clause holding struck down an owner-occupancy requirement closely matching Millbrook's. The agent must distinguish the adverse takings holding on license-vs-fee-simple grounds and mine the favorable holding the Town's attorney quietly omitted.
\item \textbf{Penn Central three-factor framework recovery.} The instructions only hint at "regulatory takings framework." The agent must independently invoke Penn Central Transportation Co. v. New York City and apply the three-factor test (economic impact, investment-backed expectations, character of government action) to each of the six properties.
\item \textbf{Per-property financial modeling across five database tables.} Settlement valuation requires joining str\_income, ltr\_comparables, settlement\_offers, fine\_accruals, and properties to compute transitional STR income at the 120-night cap, LTR conversion income, and fine exposure of \$500/day from Nov 29 through Jan 31 -- across six properties with different ownership and zoning.
\item \textbf{R-1 equal protection angle on Property 4.} 77 Seaside Terrace is excluded from the transitional permits but had a valid prior STR permit visible only in the str\_permits table. The agent must recognize that singling out one property creates an arbitrary classification raising equal protection concerns and counter-propose either inclusion or a binding rezoning commitment.
\item \textbf{Amortization vulnerability across jurisdictions.} The Town cites AVR, Inc. and City of Los Angeles v. Gage, but those involved 1-year and 5-year periods for adult entertainment and industrial uses. The agent must build the argument that 180 days with no grandfathering is constitutionally inadequate for residential STRs carrying mortgages.
\item \textbf{Dismissal-with-prejudice and tolling traps.} The settlement letter buries dismissal-with-prejudice language that would extinguish the dormant Commerce Clause and takings claims permanently. The agent must flag this and demand a mutual tolling agreement so the 1983 limitations clock does not expire during negotiations.
\end{itemize}
\vspace{0.2em}
\textbf{Files requiring search.} External legal references the agent must locate to evaluate the Town's position and ground the counter-proposal.
\begin{itemize}[leftmargin=1.25em,itemsep=0.22em,topsep=0.2em]
\item \texttt{hignell\_stark\_v\_new\_orleans\_5thcir\_2022.pdf}: Distinguish the adverse takings holding and mine the favorable dormant Commerce Clause holding against owner-occupancy rules; populates the New Orleans column of the comparative table. \emph{Source:} ca5.uscourts.gov (\url{https://www.ca5.uscourts.gov/opinions/pub/21/21-30643-CV0.pdf}).
\item \texttt{penn\_central\_takings\_test.pdf}: Apply the Penn Central three-factor regulatory takings test to assess litigation strength and ground counter-proposal terms. \emph{Source:} pennstatelawreview.org (\url{https://www.pennstatelawreview.org/footnotes/3996/}).
\end{itemize}
\end{taskcardreadablebox}

%% file: appendix-task-examples/web-admin-task1-rubrics.tex
\begin{taskrubricreadablebox}[title={Rubrics (8 checks, 50 total points)}]
\begin{enumerate}[leftmargin=1.35em,itemsep=0.55em,topsep=0.25em]
\item \textbf{8 pts.} Does the findings\_summary.csv reconstruct the full kill chain progression for 203.0.113.42 showing the correct sequence: initial probing (SQL injection at 08:13:01-05Z), credential access (brute force at 08:13:08-16Z), enumeration (IDOR user scraping at 08:16:00-09Z), exploitation (XSS at 08:19:00-08Z), data exfiltration (path traversal + config download at 08:21:00-04Z), and post-compromise actions (DELETE/PUT at 08:25:00-05Z)?
\begin{itemize}[leftmargin=1.1em,itemsep=0.12em,topsep=0.15em]
\item In findings\_summary.csv, the kill chain must show initial SQL injection probing at 08:13:01-05Z (reconnaissance/weaponization) followed by brute force authentication at 08:13:08-16Z (credential access), correctly sequenced and labeled.
\item In findings\_summary.csv, IDOR enumeration of /api/users/1 through /api/users/10 at 08:16:00-09Z must be mapped to a discovery/enumeration phase.
\item In findings\_summary.csv, XSS injection attempts at 08:19:00Z, 08:19:05Z, and 08:19:08Z using script tags, onerror handlers, and SVG onload, plus a stored XSS review submission at 08:19:02Z, must be mapped to an exploitation phase.
\item In findings\_summary.csv, path traversal and successful config file download at 08:21:00-04Z, including the critical /api/download?file=/etc/nginx/nginx.conf returning 200 with 2840 bytes, must be mapped to an exfiltration phase.
\item In findings\_summary.csv, destructive actions at 08:25:00-05Z including DELETE of products 1042-1044, PUT to /api/users/1-2, and PUT to /api/config/settings must be mapped to an impact/post-compromise phase.
\end{itemize}
\item \textbf{6 pts.} Does the cr\_compliance\_audit.csv include risk assessments that correctly connect non-compliance to specific observed attack outcomes -- particularly that the missing dotfile blocking (1c) enabled .env and .git/config exposure, the missing debug endpoint removal (1f) enabled information leakage of 8921 bytes, and the missing admin IP restriction (1e) left the admin panel accessible to external attackers?
\begin{itemize}[leftmargin=1.1em,itemsep=0.12em,topsep=0.15em]
\item Item 1c's risk description must reference that the Nikto scanner (198.51.100.77) successfully retrieved .env (487 bytes of configuration/credentials) and .git/config (312 bytes of repository metadata) due to missing dotfile blocking.
\item Item 1f's risk description must note that /api/v1/debug was accessed and returned 8921 bytes of debug information.
\item Item 1e's risk description must note that the /admin/ path was accessible to external IPs because Nginx had no source IP restriction, with the 403 coming only from the application itself.
\item In cr\_compliance\_audit.csv, firewall items (2a-2d) must connect non-compliance to the fact that internal services are directly reachable from the internet, violating the approved zone segmentation in network\_topology.json.
\end{itemize}
\item \textbf{6 pts.} Does the hardened iptables configuration (iptables\_hardened.rules) use zone-based source IP restrictions matching network\_topology.json -- specifically restricting port 3000 to 10.100.0.0/24 (DMZ), port 3001 to 10.200.0.0/24 (Management), ports 5432 and 6379 to 10.100.1.0/24 (Application zone), and SSH to 10.200.0.0/24?
\begin{itemize}[leftmargin=1.1em,itemsep=0.12em,topsep=0.15em]
\item Port 3000 (Node.js) must only accept connections from DMZ zone 10.100.0.0/24, and port 3001 (admin panel) must only accept from Management zone 10.200.0.0/24.
\item Port 5432 (PostgreSQL) and port 6379 (Redis) must only accept connections from Application zone 10.100.1.0/24.
\item SSH must be restricted to Management zone 10.200.0.0/24 and include rate limiting (e.g., '-m recent' or '-m limit') enforcing a maximum of 3 new connections per minute as specified in CR item 2e.
\item The rules must NOT use 'ACCEPT' from anywhere for any of these ports, and must NOT rely solely on localhost/127.0.0.1 bindings.
\end{itemize}
\item \textbf{6 pts.} Does the TLS posture assessment correctly identify all four certificate issues: (1) shopvault.example.com and api.shopvault.example.com using TLSv1.0 minimum instead of TLSv1.2, (2) admin.shopvault-internal.com certificate expired on 2025-11-01, (3) staging.shopvault.example.com using a self-signed certificate with 1024-bit key on SSLv3, and (4) the nginx.conf having no TLS/SSL configuration at all (no port 443 listener)?
\begin{itemize}[leftmargin=1.1em,itemsep=0.12em,topsep=0.15em]
\item The TLS assessment must identify that shopvault.example.com and api.shopvault.example.com both have protocol\_min set to TLSv1.0, violating CR item 3c which requires minimum TLSv1.2.
\item The assessment must identify that admin.shopvault-internal.com certificate expired on 2025-11-01 (9 days before the incident on Nov 10), violating CR item 3a.
\item The assessment must identify that staging.shopvault.example.com has a self-signed certificate with only 1024-bit key size and SSLv3 protocol minimum, and should have been decommissioned per CR item 3b.
\item The assessment must note that nginx.conf only has a port 80 listener with no SSL/TLS server block, meaning HTTPS is not configured at the web server level regardless of certificate availability.
\item The assessment must note that payments.shopvault.example.com (4096-bit, TLSv1.2) is the only properly configured certificate.
\end{itemize}
\item \textbf{6 pts.} Does the incident report's executive summary explicitly explain how the unimplemented change request CR-2025-0347 directly enabled the observed attacks -- specifically linking missing dotfile protection to .env/.git exposure, missing rate limiting to successful brute force, missing admin restrictions to admin panel accessibility, and missing firewall zone segmentation to direct service exposure?
\begin{itemize}[leftmargin=1.1em,itemsep=0.12em,topsep=0.15em]
\item The executive summary must state that CR item 1c (dotfile blocking) was not implemented, directly allowing 198.51.100.77 to download .env (containing credentials/secrets) and .git/config (containing repository metadata).
\item The summary must state that the absence of rate limiting (CR item 2e and web-layer) allowed 203.0.113.42 to perform 18+ rapid brute force login attempts without throttling, ultimately succeeding.
\item The summary must state that CR item 1e (/admin/ IP restriction) was not implemented, leaving the admin panel accessible from any IP, and CR items 2a-2d (firewall port restrictions) were not implemented, meaning ports 3000, 3001, 5432, and 6379 are directly accessible from the internet.
\item The summary must explicitly state that the CR was marked 'Closed (Implemented)' without verification, and this process failure enabled the entire attack surface.
\end{itemize}
\item \textbf{6 pts.} Does the incident report contain a properly populated incident timeline table with the correct chronological sequence of events, covering all three malicious IPs (203.0.113.42, 198.51.100.77, 45.33.32.156) and their respective attack categories?
\begin{itemize}[leftmargin=1.1em,itemsep=0.12em,topsep=0.15em]
\item The timeline must include 203.0.113.42's attack progression: SQL injection at 08:13:01Z, brute force starting at 08:13:08Z with success at 08:13:16Z, IDOR enumeration at 08:16:00Z, XSS attempts at 08:19:00Z, path traversal/config exfiltration at 08:21:00Z, second brute force wave at 08:23:00Z, and post-compromise destructive actions (DELETE, PUT) at 08:25:00Z.
\item The timeline must include 198.51.100.77's Nikto scanning starting at 08:14:00Z.
\item The timeline must include 45.33.32.156's Nmap scanning starting at 08:17:00Z and Shellshock attempts at 08:28:00Z.
\item All events must be in chronological order with correct timestamps, source IPs, and attack categories.
\end{itemize}
\item \textbf{6 pts.} Does the incident report's vulnerabilities section correctly classify exploited vulnerabilities using OWASP categories, distinguishing between successful exploitations (brute force login, IDOR user enumeration, .env/.git exposure, config file download, XSS reflected, post-auth destructive actions) and merely attempted attacks (SQL injection returning 500, Shellshock against non-existent CGI paths)?
\begin{itemize}[leftmargin=1.1em,itemsep=0.12em,topsep=0.15em]
\item SQL Injection (A03:2021-Injection) must be classified as attempted but failed (500 status codes), and Shellshock (A03:2021-Injection) must be classified as attempted but failed (404, CGI paths don't exist).
\item Brute Force (A07:2021-Identification and Authentication Failures) must be classified as successful (attacker gained authenticated session), and IDOR/Broken Access Control (A01:2021-Broken Access Control) must be classified as successful (10 user records enumerated with 200 status).
\item XSS (A03:2021-Injection) must be classified as successful (requests returned 200 with 15230-byte response bodies), and Path Traversal (A01:2021-Broken Access Control) must be classified as partially successful (../../etc/passwd failed with 500, but /api/download?file=/etc/nginx/nginx.conf succeeded with 200/2840 bytes).
\item Sensitive File Exposure (A05:2021-Security Misconfiguration) must be classified as successful for .env and .git/config, and the distinction between attempted and successful exploitation must be clearly made for each vulnerability.
\end{itemize}
\item \textbf{6 pts.} Does the incident report include a comprehensive remediation actions table with prioritized items covering immediate containment (block attacker IPs, revoke compromised sessions), short-term fixes (deploy hardened configs, renew expired certs), and long-term improvements (WAF deployment, rate limiting, automated CR verification)?
\begin{itemize}[leftmargin=1.1em,itemsep=0.12em,topsep=0.15em]
\item P1/Immediate actions must include: block malicious IPs (203.0.113.42, 198.51.100.77, 45.33.32.156) at the firewall, revoke all active sessions (especially the brute-forced session), reset compromised user credentials for at least users 1 and 2, and verify/restore deleted products (1042-1044) and revert the config settings change.
\item P2/Short-term actions must include: deploy the hardened nginx.conf and iptables.rules, renew the expired admin.shopvault-internal.com certificate, decommission staging.shopvault.example.com, and enforce TLS 1.2 minimum.
\item P3/Long-term actions must include: deploy a Web Application Firewall (WAF), implement rate limiting at application and infrastructure layers, establish automated post-implementation verification for change requests, and enhance logging and monitoring capabilities.
\item All remediation actions must be specific to the findings from this incident, not generic security recommendations.
\end{itemize}
\end{enumerate}
\end{taskrubricreadablebox}

%% file: appendix-task-examples/web-admin-task1-task-card.tex
\begin{taskcardreadablebox}[title={Task Card}]
\textbf{Summary.} This task asks the agent to act as the web administrator investigating ShopVault's November 10, 2025 incident -- change request CR-2025-0347 was closed without verification -- by reconstructing attack chains from access logs against threat intel, auditing every CR item against live nginx and iptables configurations, hardening both configs against the actual zone topology, and populating a NIST-structured incident report.\par
\vspace{0.35em}
\begin{tabularx}{\textwidth}{@{}>{\bfseries}p{0.23\textwidth}Y@{}}
ONET task & Monitor systems for intrusions or denial of service attacks, and report security breaches to appropriate personnel. \\
Expert desire & 3.80 (scale: 0-5, from WorkBank) \\
In practice & correlating access\_logs.csv against threat\_indicators.json to label each event with a kill-chain phase and success flag, walking every CR-2025-0347 item against the running nginx.conf and iptables.rules to find the gap that enabled each observed exploit, and producing hardened replacement configs that respect the network\_topology.json zone CIDRs. \\
\end{tabularx}
\vspace{0.35em}
\textbf{Why desired.} Incident reconstruction is an evidence-joining job -- match an IP across hundreds of log lines, line up the matching IOC, trace it to the configuration weakness, then write it up. Doing this by hand under post-incident pressure is slow and error-prone. An agent that produces the audited matrix and hardened configs lets the administrator focus on containment decisions and executive communication.\par
\vspace{0.35em}
\textbf{Reasoning challenges.}
\begin{itemize}[leftmargin=1.25em,itemsep=0.22em,topsep=0.2em]
\item \textbf{Causal chain from CR non-compliance to attack success.} CR item 1c (dotfile blocking) was never implemented, so nginx.conf still allows .env access and Nikto scanner 198.51.100.77 retrieved .env (HTTP 200, 487 bytes). Each compliance gap in cr\_compliance\_audit.csv must be tied to specific log entries showing exploitation.
\item \textbf{Zone segmentation reconciliation for hardened iptables.} network\_topology.json CIDRs must drive source-IP restrictions: port 3000 to DMZ (10.100.0.0/24), port 3001 to Management (10.200.0.0/24), ports 5432/6379 to Application (10.100.1.0/24), SSH to Management (10.200.0.0/24).
\item \textbf{Adapting the WordPress-Nginx reference, not copying it.} The external iptables\_reference.rules contains FTP rules on port 21 that are not part of ShopVault's approved architecture. The agent must exclude these rather than blindly inherit them.
\item \textbf{Multi-actor kill chain differentiation.} Three malicious IPs (203.0.113.42, 198.51.100.77, 45.33.32.156) and three legitimate users are interleaved in the logs. Each threat actor's kill chain must be reconstructed independently while a unified chronological timeline is maintained for the incident report.
\item \textbf{TLS gap as absence, not misconfiguration.} Cross-referencing ssl\_cert\_audit.csv against CR items 3a-3c against nginx.conf reveals that nginx has no port 443 listener and no SSL/TLS directives at all -- HTTPS is entirely unconfigured despite valid certificates existing.
\item \textbf{SSH rate-limit threshold extraction.} CR item 2e specifies SSH rate limiting at 3 connections per minute (embedded in the change request text) -- the hardened iptables must implement this exactly using \texttt{-m recent} or \texttt{-m limit}, not generic rate-limiting.
\item \textbf{Kill chain phase taxonomy applied independently.} The instructions name "kill chain phase" without defining the phases -- the agent must apply Cyber Kill Chain or MITRE ATT\&CK labels (reconnaissance, credential access, enumeration, exploitation, exfiltration, post-compromise/impact) on its own.
\item \textbf{NIST SP 800-61 report structure inferred from template.} The incident\_report\_template.html follows NIST SP 800-61 Rev. 2 (preparation, detection, containment, eradication, recovery, lessons learned) -- the agent must recognize the structure and populate sections accordingly, with the executive summary tracing causation from unverified CR items to observed attack outcomes.
\end{itemize}
\vspace{0.2em}
\textbf{Files requiring search.} External firewall reference the agent must locate as a structural baseline for the hardened iptables ruleset.
\begin{itemize}[leftmargin=1.25em,itemsep=0.22em,topsep=0.2em]
\item \texttt{iptables\_reference.rules}: WordPress-Nginx iptables hardening reference providing default-deny baseline and service exposure patterns; must be adapted to drop FTP and incorporate ShopVault's zone-based architecture. \emph{Source:} github.com/abdusfauzi/wordpress-nginx (\url{https://raw.githubusercontent.com/abdusfauzi/wordpress-nginx/master/etc/iptables.firewall.rules}).
\end{itemize}
\end{taskcardreadablebox}

%% file: sections/appendix_prompts.tex
\section{Prompt Templates and Evaluation Interface}
\label{app:prompts}

In this appendix, we record the runtime prompts used by the JobBench evaluation. Section~\ref{app:prompts:agent} reports the agent evaluation prompt used by all CLI runners. Section~\ref{app:prompts:judge} reports the rubric-level judge prompt used for automatic evaluation.

\subsection{Agent Evaluation Prompt}
\label{app:prompts:agent}

The OpenCode, Claude Code, and Codex CLI runners share a single task prompt. Before each evaluation, the runner copies the task into an isolated temporary workspace and substitutes the path placeholders shown below. The prompt directs the agent to the task instructions, the reference files, and the output directory.

\begin{promptlisting}[title=Agent Evaluation Prompt]
=== TASK FOLDER ===
{temp_task_folder}

=== INSTRUCTIONS ===
1. Read the TASK_INSTRUCTIONS.txt file in the task folder above
2. Based on the Reference Files section in TASK_INSTRUCTIONS.txt, read the corresponding files from the same task folder using appropriate tools.
3. Complete the task as specified in TASK_INSTRUCTIONS.txt
4. Only save the final deliverables to the output directory specified below. Do not save any intermediate or temporary files.

=== OUTPUT DIRECTORY ===
{temp_output_dir}

IMPORTANT:
- All reference files are in the task folder: {temp_task_folder}
- Only save the final deliverables to the output directory {temp_output_dir}. Do not save any intermediate or temporary files.
- You MUST only access files within {temp_workspace} or search online for new reference files if you find needed. Do NOT access any files or directories in this system outside of this path.
- If you encounter ambiguous or conflicting information, analyze the conflict, explain your reasoning, and justify the approach you choose.
- If a file cannot be read directly (e.g., .xlsx, .docx, .db, .pptx), use appropriate tools, MCP servers, or code to extract and process its contents.
\end{promptlisting}

\subsection{Judge Prompt}
\label{app:prompts:judge}

The judge evaluates one rubric at a time. Before each call, JobBench extracts text from the model's output directory and normalizes the rubric criteria. The prompt then supplies the rubric, all criteria, and the extracted output contents, and requires a structured JSON response.

\begin{promptlisting}[title=Judge Prompt]
You are an evaluation judge. Your task is to evaluate ALL criteria for a single rubric.

## Rubric Description
{rubric_text}

## Criteria to Evaluate (Judge ALL of them)
{criteria_list_text}

## Output Files Content
The following are the contents of all output files to evaluate:

{file_contents}

## Evaluation Rules
- Evaluate EACH criterion listed above independently
- For each criterion: determine if it PASSES or FAILS
- Semantic matching is acceptable (you don't need exact wording match)
- Binary judgment for each criterion: PASS or FAIL only
- The rubric passes ONLY if ALL criteria pass

## Output Format
Return your judgment as a JSON object with EXACTLY this structure (no markdown, no extra text):
{
  "criteria_results": [
    {"index": 0, "passed": true/false, "reasoning": "...", "evidence": "..."},
    {"index": 1, "passed": true/false, "reasoning": "...", "evidence": "..."}
  ],
  "rubric_passed": true/false,
  "overall_reasoning": "Summary of why the rubric passed or failed"
}

IMPORTANT:
- criteria_results array must have exactly {criterion_count} items (one for each criterion)
- rubric_passed should be true ONLY if ALL criteria passed
- Include specific evidence from the output files
\end{promptlisting}

When the rubric wording requires visual evidence, the judge call additionally attaches image files from the model's output directory. The evidence bullet then appends ``and the attached images'', and the user content includes the following multimodal attachment block before the image payloads.

\begin{promptlisting}[title=Vision Attachment Block]
## Attached Images ({n} files)
Image 1: {filename_1}
Image 2: {filename_2}
...
\end{promptlisting}

%% file: sections/appendix_supplementary_analysis.tex
\section{Supplementary Analyses on JobBench's Position in the AI Labour Market}
\label{app:supplementary}

\paragraph{Worker perspective.}
JobBench starts from what workers actually want to delegate, not just from which jobs look economically exposed. That matters because demand for automation and model capability do not always line up: some tasks workers want to offload are already easy for current systems, while others still fail even when demand is high. JobBench is useful because it shows this gap at the task level instead of hiding it inside occupation averages.

\paragraph{Labour-market outlook.}
Current AI can already speed up routine work, consistent with prior evidence of 15--34\% productivity gains in customer-service settings \citep{brynjolfsson2023genai}, but it still struggles with the judgment-heavy details needed for reliable delegation. JobBench is useful for tracking whether future capability gains are expanding human productivity or increasing replacement pressure.